# ORGaNICs: A Theory of Working Memory in Brains and Machines

David J. Heeger and Wayne E. Mackey

Department of Psychology and Center for Neural Science
New York University
4 Washington Place, room 809
New York, NY 10003

Email: david.heeger@nyu.edu

## Abstract

Working memory is a cognitive process that is responsible for temporarily holding and manipulating information. Most of the empirical neuroscience research on working memory has focused on measuring sustained activity and sequential activity in prefrontal cortex and/or parietal cortex during delayed-response tasks, and most of the models of working memory have been based on neural integrators. But working memory means much more than just holding a piece of information online. We describe a new theory of working memory, based on a recurrent neural circuit that we call ORGaNICs (Oscillatory Recurrent GAted Neural Integrator Circuits). ORGaNICs are a variety of Long Short Term Memory units (LSTMs), imported from machine learning and artificial intelligence, that can be used to explain the complex dynamics of delay-period activity during a working memory task. The theory is analytically tractable so that we can characterize the dynamics, and the theory provides a means for reading out information from the dynamically varying responses at any point in time, in spite of the complex dynamics. ORGaNICs can be implemented with a biophysical (electrical circuit) model of pyramidal cells, combined with shunting inhibition via a thalamocortical loop. Although introduced as a computational theory of working memory, ORGaNICs can also be applied to sensory and motor systems, and ORGaNICs offer computational advantages compared to other varieties of LSTMs that are commonly used in AI applications. Consequently, ORGaNICs are a framework for canonical computation in brains and machines.

## Introduction

Working memory involves much more than simply holding a piece of information online. In the psychology literature, the idea of working memory includes manipulating online information dynamically in the context of new sensory input. For example, understanding a complex utterance (with multiple phrases) often involves disambiguating the syntax and/or semantics of the beginning of the utterance based on information at the end of the sentence. Consider the following two sentences:

The athlete realized his goals were unattainable.

The athlete realized his goals quickly.

The first part of each sentence is ambiguous because "his goals" might be the direct object of the verb (as in the second sentence) or the subject of a complement clause (as in the first sentence). Furthermore, one can insert any number of modifiers between the ambiguity and the point where it is resolved:

The athlete realized his goals to qualify for this year's Olympic team (quickly/were unattainable).

Or any number of relative clauses:

The athlete realized his goals which were formed during childhood (quickly/were unattainable).

Comprehending these sentences involves representing and manipulating long-term dependencies, i.e., maintaining a representation of the ambiguous information, and then changing that representation when the ambiguity is resolved.

Most of the empirical neuroscience research on working memory, by contrast, has focused only on maintenance, not manipulation, during delayed-response tasks (Fuster 1973; Fuster and Alexander 1971; Jacobsen 1935). A large body of experimental research has measured sustained activity in prefrontal cortex (PFC) and/or parietal cortex during the delay periods of various such tasks including memory-guided saccades (e.g., Constantinidis et al. 2002; Funahashi et al. 1989; Gnadt and Andersen 1988; Goldman-Rakic 1995; Schluppeck et al. 2006; Srimal and Curtis 2008) and delayed-discrimination and delayed match-to-sample tasks (e.g., Hussar and Pasternak 2012; Romo et al. 1999). Most of the models of working memory, based on neural integrators, are aimed to explain sustained delay-period activity or to explain well-established behavioral phenomena associated with sustained activity (Almeida et al. 2015; Compte et al. 2000; Wang 1999; 2001; Wimmer et al. 2014). There are, however, a variety of experimental results that are difficult to reconcile with sustained delay-period activity and neural integrator models. First, some (if not the majority of) neurons exhibit sequential activation during the delay periods such that the activity is "handed off" from one neuron to the next during a delay period with each individual neuron being active only transiently (e.g., Fujisawa et al. 2008; Harvey et al. 2012; Ikegaya et al. 2004; Schmitt et al. 2017), or they exhibit complex dynamics during the delay periods (Brody et al. 2003; Kobak et al. 2016; Machens et al. 2005; Markowitz et al. 2015; Murray et al. 2017; Shafi et al. 2007), not just constant, sustained activity. Second, complex dynamics (including oscillations) are evident also in the synchronous activity (e.g., as measured with local field potentials) of populations of neurons (Lundqvist et al. 2016; Pesaran et al. 2002). Third, some of the same neurons exhibit activity that is dependent on task demands (Gold and Shadlen 2003; Janssen and Shadlen 2005; Platt and Glimcher 1999; Watanabe 1996). Fourth, some of these neurons appear to contribute to different cognitive processes (controlling attention, working memory, decision making, motor preparation, motor control), either for different tasks or during different phases of task execution over time (Hasegawa et al. 1998; Lebedev et al. 2004; Messinger et al. 2009; Quintana and Fuster 1992; Rowe et al. 2000).

Long Short Term Memory units (LSTMS, Hochreiter and Schmidhuber 1997) are machine learning (ML) / artificial intelligence (AI) algorithms  that are capable of representing and manipulating long-term dependencies, in a manner that is analogous to the concept of working memory in the psychology literature. LSTMs are a class of recurrent neural networks (RNNs). A number of variants of the basic LSTM architecture have been developed and tested for a variety of AI applications. One of the variants, called a gated recurrent unit (GRU), is currently popular (Cho et al. 2014). For a nice introduction to RNNs and LSTMs, see this blog by Christopher Olah:

http://colah.github.io/posts/2015-08-Understanding-LSTMs/.

And see this blog by Andrej Karpathy for some demonstrations of AI applications that use RNNs and LSTMs:

http://karpathy.github.io/2015/05/21/rnn-effectiveness/

LSTMs have been used for a variety of machine learning problems including language modeling, neural machine translation, and speech recognition (Assael et al. 2016; Chan et al. 2016; Cho et al. 2014; Chorowski et al. 2015; Graves 2013; Graves et al. 2013; Graves et al. 2014; Sutskever et al. 2014; van den Oord et al. 2016). In these and other tasks, the input stimuli contain information across multiple timescales, but the ongoing presentation of stimuli makes it difficult to correctly combine that information over time (Hochreiter et al. 2001; Pascanu et al. 2013). An LSTM handles this problem by updating its internal state over time with a pair of "gates": the update gate selects which part(s) of the current input to combine with the current state, and the reset gate selectively deletes part(s) of the current state. The gates are themselves computed from the current inputs and the current outputs. This enables LSTMs to maintain and manipulate a representation of some of the inputs, until needed, without interference from other inputs that come later in time.

Here, we describe a neurobiological theory of working memory, based on a recurrent neural circuit that we call ORGaNICs (Oscillatory Recurrent GAted Neural Integrator Circuits). The theory is an extension of Heeger's Theory of Cortical Function (Heeger 2017), although various complementary approaches have each achieved some of the same goals (Costa et al. 2017; Goldman 2009; Lundqvist et al. 2010; Lundqvist et al. 2011; O'Reilly and Frank 2006). ORGaNICs are a variety of LSTMs. Having the capability of an LSTM ensures that the ORGaNICs can solve relatively complicated tasks (much more complicated than the typical delayed-response tasks used in most cognitive psychology and neuroscience experiments). ORGaNICs can exhibit complex dynamics (by using complex-valued weights and responses), including sequential activation, but the theory provides a means for "reading out" information from the dynamically varying responses at any point in time, in spite of the complex dynamics. ORGaNICs can be implemented with a simplified biophysical (equivalent electrical circuit) model of pyramidal cells with separate compartments for the soma, apical dendrite, and basal dendrite. Although introduced as a computational theory of working memory, ORGaNICs are also applicable to models of sensory processing (because deep nets are a special case of stacked ORGaNICs), motor preparation and motor control (because ORGaNICs may be used to generate signals with complex dynamics, again by using complex-valued weights and responses), and time-series prediction (again because the complex-valued weights may be used to generate signals with complex dynamics; the weights correspond to a collection of predictive basis functions, damped oscillators of various frequencies). ORGaNICs offer computational advantages compared to other varieties of LSTMs that are commonly used in AI applications. Consequently, ORGaNICs are a framework for canonical computation in brains and machines.

**ORGaNICs**

The responses of a population of PFC (or parietal) neurons are represented by a vector $\mathbf{y} = (y_1, y_2, \ldots, y_j, \ldots, y_N)$ where the subscript $j$ indexes different neurons in the population. Note that we use boldface lowercase letters to represent vectors and boldface uppercase to denote matrices. The responses $\mathbf{y}$ depend on an input drive $\mathbf{z}$ and a recurrent drive $\hat{\mathbf{y}}$. The responses $\mathbf{y}$ are also modulated by two other populations of neurons: $\boldsymbol{\alpha}$ and $\mathbf{b}$. The variables ($\mathbf{y}$, $\hat{\mathbf{y}}$, $\mathbf{z}$, $\boldsymbol{\alpha}$, and $\mathbf{b}$) are each functions of time, e.g., $\mathbf{y}(t)$, but we drop the explicit dependence on $t$ to simplify the notation except when it is helpful to disambiguate time steps.

[A note about terminology. We use the term "recurrent", following the AI and machine literature, to mean feedback processing over time, but the term "recursive" is typically used instead of "recurrent" in the signal processing literature (e.g., a recursive filter with an infinite impulse response).]

The starting point is the hypothesis that neural responses minimize an energy function that represents a compromise between between an input drive and recurrent drive, over time:

$$E = \tfrac{1}{2} \int_t \sum_j \left( \frac{b_j^+}{1+b_j^+} \right) \left[ y_j - z_j \right]^2 + \left( \frac{1}{1+b_j^+} \right) \left[ y_j - \left( \frac{1}{1+\alpha_j^+} \right) \hat{y}_j \right]^2$$

$$\propto \tfrac{1}{2} \sum_t \sum_j \left( \frac{b_j^+}{1+b_j^+} \right) \left[ y_j - z_j \right]^2 + \left( \frac{1}{1+b_j^+} \right) \left[ y_j - \left( \frac{1}{1+\alpha_j^+} \right) \hat{y}_j \right]^2$$

(1)

$$\alpha_j^+ \geq 0 \ \text{ and } \ b_j^+ \geq 0 \ ,$$

where the superscript + is a rectifying output nonlinearity. Halfwave rectification is the simplest form of this rectifying nonlinearity, but other output nonlinearities could be substituted, e.g., sigmoid, exponentiation, half-squaring (Heeger 1992a), normalization (Carandini and Heeger 2012; Heeger 1992b), etc. The second line of **Eq. 1** is obtained by discretely sampling time, and the proportionality constant is equal to the time step $\Delta t$.

**Figure 1. ORGaNIC architecture.** Diagram of connections in an example ORGaNIC. Solid lines/curves are excitatory and dashed curves are inhibitory. Only a few of the recurrent connections are shown to minimize clutter. Modulatory connections not shown.

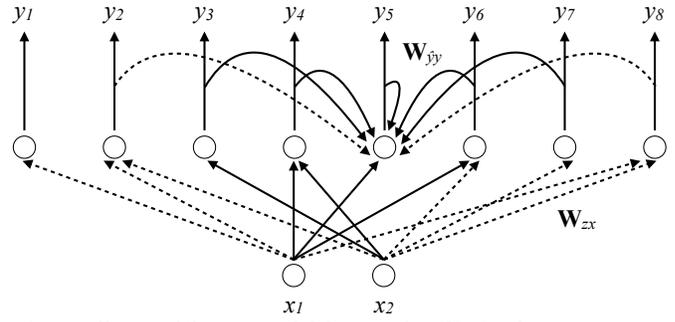

The neural responses are modeled as dynamical processes that minimize the energy $E$ over time. Taking derivatives of **Eq. 1**:

$$\tau_y \frac{dy_j}{dt} = -\frac{dE}{dy_j} \tag{2}$$

$$= -\left(\frac{b_j^+}{1+b_j^+}\right)\left[y_j - z_j\right] - \left(\frac{1}{1+b_j^+}\right)\left[y_j - \left(\frac{1}{1+\alpha_j^+}\right)\hat{y}_j\right]$$

$$= -\left(\frac{b_j^+}{1+b_j^+}\right)y_j + \left(\frac{b_j^+}{1+b_j^+}\right)z_j - \left[1-\left(\frac{b_j^+}{1+b_j^+}\right)\right]y_j + \left(\frac{1}{1+b_j^+}\right)\left(\frac{1}{1+\alpha_j^+}\right)\hat{y}_j \quad,$$

$$= -y_j + \left(\frac{b_j^+}{1+b_j^+}\right)z_j + \left(\frac{1}{1+b_j^+}\right)\left(\frac{1}{1+\alpha_j^+}\right)\hat{y}_j$$

where $\tau_y$ is the intrinsic time-constant of the neurons. We note the recurrent drive $\hat{y}_j(t)$ depends on the responses $y_j(t-\Delta t)$ from an instant earlier in time, and that the gradient $dE/dy_j$ is with respect to $y_j(t)$, for a specific time $t$, so we do not apply the chain rule to $\hat{y}_j(t)$.

The input drive depends on a weighted sum of the inputs:

$$\mathbf{z} = \mathbf{W}_{zx}\mathbf{x} + \mathbf{c}_z \quad, \tag{3}$$

where $\mathbf{x} = (x_1, x_2,\ldots, x_j,\ldots, x_M)$ is a vector representing the time-varying responses of a population of input neurons. These input neurons may also be in PFC, perhaps in the input layer. The encoding matrix (also called the embedding matrix) $\mathbf{W}_{zx}$ is an $N\mathrm{x}M$ matrix of weights and $\mathbf{c}_z$ is an $N$-vector of additive offsets. You can think of the rows of $\mathbf{W}_{zx}$ as being the "receptive fields" of the neurons, and $\mathbf{c}_z$ as determining the spontaneous firing rates.

The recurrent drive depends on a weighted sum of the responses:

$$\hat{\mathbf{y}} = \mathbf{W}_{\hat{y}y}\mathbf{y} + \mathbf{c}_{\hat{y}} \quad. \tag{4}$$

The recurrent weight matrix $\mathbf{W}_{\hat{y}y}$ is an $N\mathrm{x}N$ matrix and $\mathbf{c}_{\hat{y}}$ an $N$-vector of additive offsets. If $\mathbf{W}_{\hat{y}y}$ is the identity matrix, then each neuron receives a recurrent excitatory connection from itself. If $\mathbf{W}_{\hat{y}y}$ has a diagonal structure, then each neuron receives recurrent connections from itself and its neighbors. This could, for example, be a center-surround architecture in which the closest recurrent connections are excitatory and the more distant ones are inhibitory.

The readout is also a weighted sum of the responses:

$$\mathbf{W}_{ry}\mathbf{y} + \mathbf{c}_r \ . \tag{5}$$

where $\mathbf{W}_{ry}$ is the readout weight matrix (also called the decoding matrix). If $\mathbf{W}_{ry}$ is the identity matrix, then the readout is the same as the output responses $\mathbf{y}$. The readout may (optionally) be followed by an output nonlinearity, e.g., halfwave rectification, sigmoid, hyperbolic tangent, normalization (Carandini and Heeger 2012; Heeger 1992b), etc.

The modulators, $\boldsymbol{\alpha}$ and $\mathbf{b}$, are analogous to the reset gates and update gates, respectively, in a GRU (Cho et al. 2014). The time-varying value of each $b_j$ determines the effective time-constant of the corresponding response time-course $y_j$. The first term of $E$ (**Eq. 1**) drives the output responses $\mathbf{y}$ to match the input drive $\mathbf{x}$, and the second term drives the output responses to match the recurrent drive $\hat{\mathbf{y}}$. Consequently, if $b_j$ is large then the response time-course $y_j$ is dominated by the input drive, and if $b_j$ is small then the response time-course is dominated by the recurrent drive. The time-varying value of $\alpha_j$ determines the gain of the recurrent drive $\hat{y}_j$. If $\alpha_j$ is large then the recurrent drive is shut down regardless of the value of $b_j$.

A (leaky) neural integrator corresponds to the special case in which $\alpha_j = 0$, $b_j = b$ is the same for all neurons $j$ and constant over time, and $\mathbf{c}_z = \mathbf{c}_{\hat{y}} = \mathbf{0}$. For this special case, we can rewrite **Eq. 2**:

$$\tau_y \frac{d\mathbf{y}}{dt} = -\mathbf{y} + \lambda \mathbf{z} + (1-\lambda)\hat{\mathbf{y}} \tag{6}$$

$$\hat{\mathbf{y}} = \mathbf{W}_{\hat{y}y}\mathbf{y}$$

$$\lambda = \left(\frac{|b|}{1+|b|}\right) \text{ and } (1-\lambda) = \left(\frac{1}{1+|b|}\right),$$

where $0 \leq \lambda \leq 1$. Even simpler is when $\mathbf{W}_{\hat{y}y} = \mathbf{I}$ (where $\mathbf{I}$ is the identity matrix):

$$\tau_y \frac{dy_j}{dt} = -y_j + \lambda z_j + (1-\lambda)y_j = \lambda\left(z_j - y_j\right) \tag{7}$$

i.e.,

$$\tau'_y \frac{dy_j}{dt} = -y_j + z_j \tag{8}$$

$$\tau'_y = \frac{\tau_y}{\lambda} \ .$$

where $\tau_y$ is the intrinsic time-constant and $\tau'_y$ is the effective time-constant. For this simple special case, each neuron acts like a shift-invariant linear system, i.e., a recursive linear filter with an exponential impulse response function. If the input drive $z_j$ is constant over time, then the responses $y_j$ exhibit an exponential time course with steady state $y_j = z_j$, and time constant $\tau'_y$. This special case reveals how $\lambda$, and consequently $b$ determines the effective time-constant of the leaky integrator.

We introduce a change of variables (from $\boldsymbol{\alpha}$ to $\mathbf{a}$) in **Eq. 2**:

$$\tau_y \frac{dy_j}{dt} = -y_j + \left(\frac{b_j^+}{1+b_j^+}\right)z_j + \left(\frac{1}{1+a_j^+}\right)\hat{y}_j \ . \tag{9}$$

For **Eqs. 2** and **9** to be identical:

$$\left(1 + a_j^+\right) = \left(1 + b_j^+\right)\left(1 + \alpha_j^+\right) \tag{10}$$

i.e., $a_j^+ = \alpha_j^+ b_j^+ + \alpha_j^+ + b_j^+$ ,

but we will not enforce this constraint, allowing $a_j$ to take on any value. A leaky integrator now corresponds to the special case in which $a_j = b_j$. The original formulation of **Eq. 2** is more elegant because the roles of $\alpha_j$ and $b_j$ are distinct: $\alpha_j$ determines the gain of the recurrent drive and $b_j$ determines the effective time-constant. But the formulation of **Eq. 9** is more convenient because it has separate modulators, $a_j$ and $b_j$, for the input drive and the recurrent drive, respectively. So we use the formulation of **Eq. 9** in what follows.

By analogy with LSTMs, the modulators **a** and **b** are themselves modeled as dynamical systems that depend on weighted sums of the inputs and outputs:

$$\tau_a \frac{d\mathbf{a}}{dt} = -\mathbf{a} + \mathbf{W}_{ax}\mathbf{x} + \mathbf{W}_{ay}\mathbf{y} + \mathbf{c}_a \tag{11}$$

$$\tau_b \frac{d\mathbf{b}}{dt} = -\mathbf{b} + \mathbf{W}_{bx}\mathbf{x} + \mathbf{W}_{by}\mathbf{y} + \mathbf{c}_b \quad . \tag{12}$$

The weights in the various weight matrices ($\mathbf{W}_{zx}$, $\mathbf{W}_{\hat{y}y}$, $\mathbf{W}_{ry}$, $\mathbf{W}_{ax}$, $\mathbf{W}_{ay}$, $\mathbf{W}_{bx}$, $\mathbf{W}_{by}$) and the offset vectors ($\mathbf{c}_z$, $\mathbf{c}_{\hat{y}}$, $\mathbf{c}_r$, $\mathbf{c}_a$, $\mathbf{c}_b$) are presumed to be learned, depending on task demands. But the elements of these weight matrices and offset vectors were prespecified (not learned), fixed values for each of the simulations in this paper.

In summary, $\mathbf{x}$ is a time-varying vector of inputs and $\mathbf{y}$ is a time-varying vector of output responses. The output responses depend on a weighted sum of the inputs, and a recurrent weighted sum their own responses (an example is depicted in **Fig. 1)**. The output responses are also modulated by two time-varying modulators, $\mathbf{a}$ and $\mathbf{b}$, which determine the effective time-constant and the recurrent gain. Each of these modulators depends on a weighted sum of the inputs and outputs. So there are two nested recurrent circuits. First, the responses $\mathbf{y}$ depend on the recurrent drive ($\hat{y}$) which depends on a weighted sum of the responses. Second the responses are modulated by a pair of modulators ($\mathbf{a}$ and $\mathbf{b}$), each of which depends on a weighted sum of the responses.

The figures in the following subsections illustrate some of the different operating regimes of ORGaNICs. To briefly preview some of the key results… When $a_j = b_j > 0$, the $j$th neuron behaves like a leaky integrator, i.e., a recursive linear filter with an exponential impulse response function. When $a_j = b_j = 0$, the response of the $j$th neuron is determined entirely by the recurrent drive and the circuit may exhibit sustained delay-period activity and/or ongoing, stable oscillations. When $a_j > b_j$, the recurrent drive may be shut off (i.e., the neuron's response is reset). When $b_j > 0$ and $a_j = 0$, the neuron behaves like a full (not leaky) integrator. When the eigenvectors and eigenvalues of the recurrent weight matrix are composed of complex values, the responses may exhibit oscillations. Each eigenvector of the recurrent weight matrix is associated with a basis function, a pattern of activity across the population of neurons and over time. The basis functions are damped oscillators when $\mathbf{a} = \mathbf{b}$. All this is explained in detail through the examples that follow.

**Implementation**

The ORGaNIC algorithm is expressed by the following system of discrete-time equations, looping over $t$ in increments of $\Delta t$ :

$$\mathbf{z}(t) = \mathbf{W}_{zx}\mathbf{x}(t) + \mathbf{c}_z$$

$$\hat{\mathbf{y}} = \mathbf{W}_{\hat{y}y}\mathbf{y} + \mathbf{c}_{\hat{y}}$$

$$\Delta\mathbf{a}(t) = \frac{\Delta t}{\tau_a}\Big[-\mathbf{a}(t) + \mathbf{W}_{ax}\mathbf{x}(t) + \mathbf{W}_{ay}\mathbf{y}(t) + \mathbf{c}_a\Big]$$

$$\mathbf{a}(t+\Delta t) = \mathbf{a}(t) + \Delta\mathbf{a}(t)$$

$$\Delta\mathbf{b}(t) = \frac{\Delta t}{\tau_b}\Big[-\mathbf{b}(t) + \mathbf{W}_{bx}\mathbf{x}(t) + \mathbf{W}_{by}\mathbf{y}(t) + \mathbf{c}_b\Big] \qquad . \qquad (13)$$

$$\mathbf{b}(t+\Delta t) = \mathbf{b}(t) + \Delta\mathbf{b}(t)$$

$$\Delta y_j(t) = \frac{\Delta t}{\tau_y}\Big[-y_j(t) + \Big(\frac{b_j^+(t)}{1+b_j^+(t)}\Big)z_j(t) + \Big(\frac{1}{1+a_j^+(t)}\Big)\hat{y}_j(t)\Big]$$

$$\mathbf{y}(t+\Delta t) = \mathbf{y}(t) + \Delta\mathbf{y}(t)$$

The algorithm is incremental, meaning that it stores only a vector of values (one value for each neuron) for each of the variables: $\mathbf{x}$, $\mathbf{y}$, $\mathbf{z}$, $\mathbf{a}$, and $\mathbf{b}$. Each of these values is updated during each iteration of the loop from one time point to the next. (Note, however, that we often store these variables as arrays so that we can plot the results over time after finishing the loop).

The example shown in **Fig. 2**, depicts a simulation of the memory-guided saccade task, using the network architecture depicted in **Fig. 1**, with 8 neurons (i.e., each of $\mathbf{y}$, $\mathbf{z}$, $\mathbf{a}$, and $\mathbf{b}$ are 8-dimensional vectors). The input $\mathbf{x}$ consisted of 4 time courses, the first two of which represented the presentation of the two-dimensional location of a target (**Fig. 2a**; blue, horizontal position; orange, vertical position). Note that the inputs were not selective for saccade amplitude. Instead, the values of these inputs simply increased with the radial position of the target. It would be straightforward to replace these two inputs with a large number of inputs that are each selective for a different two-dimensional location in the visual field. But this 2D input is convenient, because of its much lower dimensionality, for introducing the key concepts of ORGaNICs. The input also consisted of the time-courses of two cues, one of which indicated the beginning of the trial (at time 0 ms) and the other of which indicated the end of the delay period (at time 3000 ms). The input drive (**Fig. 2b**) consisted of 8 time-courses, each of which was responsive to the polar angle location of the target. I.e., the matrix of receptive fields, $\mathbf{W}_{zx}$, was an 8x4 matrix:

$$\mathbf{W}_{zx} = \begin{pmatrix} -0.5 & 0 & 0 & 0 \\ -0.3536 & -0.3536 & 0 & 0 \\ 0 & 0.5 & 0 & 0 \\ 0.3536 & 0.3536 & 0 & 0 \\ 0.5 & 0 & 0 & 0 \\ 0.3536 & -0.3536 & 0 & 0 \\ 0 & -0.5 & 0 & 0 \\ -0.3536 & -0.3536 & 0 & 0 \end{pmatrix} \qquad (14)$$

**Figure 2. Sustained delay-period activity. a.** Input (**x**) corresponding to target presentation. Blue, horizontal position of target. Orange, vertical position. **b.** Input drive. **c, d.** Responses of the modulators, **a** and **b**. **e.** Output responses (**y**). **f.** Recurrent weight matrix ($\mathbf{W}_{\hat{y}y}$). The values of the weights range from -0.1213 to 0.3640 (white, positive weights; black, negative weights).

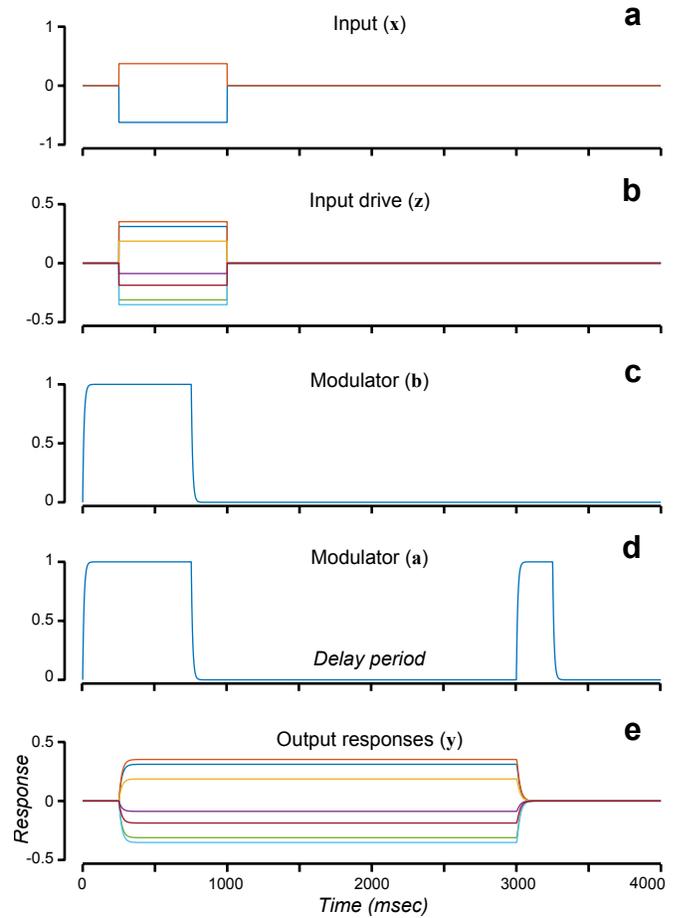

The first two columns of $\mathbf{W}_{zx}$ were, in fact, computed (for reasons explained below) as the first two eigenvectors of the recurrent weight matrix $\mathbf{W}_{\hat{y}y}$. The weight matrices for **a** and **b** were chosen to reflect the other two inputs:

$$\mathbf{W}_{ax} = \begin{pmatrix} 0 & 0 & 1 & 1 \\ 0 & 0 & 1 & 1 \\ 0 & 0 & 1 & 1 \\ 0 & 0 & 1 & 1 \\ 0 & 0 & 1 & 1 \\ 0 & 0 & 1 & 1 \\ 0 & 0 & 1 & 1 \\ 0 & 0 & 1 & 1 \end{pmatrix} \quad \mathbf{W}_{bx} = \begin{pmatrix} 0 & 0 & 1 & 0 \\ 0 & 0 & 1 & 0 \\ 0 & 0 & 1 & 0 \\ 0 & 0 & 1 & 0 \\ 0 & 0 & 1 & 0 \\ 0 & 0 & 1 & 0 \\ 0 & 0 & 1 & 0 \\ 0 & 0 & 1 & 0 \end{pmatrix} \qquad (15)$$

Consequently, the response time-courses of **a** and **b** followed the two cues (**Figs. 2c,d**). The recurrent weights $\mathbf{W}_{\hat{y}y}$ were chosen to have a center-surround architecture; each row of $\mathbf{W}_{\hat{y}y}$ had a large positive value along the diagonal (self-excitation), flanked by smaller positive values, and surrounded by small negative values (**Fig. 2f**). The other weights and offsets were all set to zero: $\mathbf{W}_{ay}=\mathbf{0}$, $\mathbf{W}_{by}=\mathbf{0}$, $\mathbf{c}_z=\mathbf{0}$, $\mathbf{c}_a=\mathbf{0}$, and $\mathbf{c}_b=\mathbf{0}$. The responses **y** followed the input drive **z** at the beginning of the simulated trial because **a** and **b** were large (=1, corresponding to a short effective time constant). The values of $\mathbf{a}$ and $\mathbf{b}$ then switched to be small (=$\mathbf{0}$, corresponding to a long effective time constant) before the target was extinguished, so the output responses $\mathbf{y}$ exhibited sustained delay-period activity. Finally, the values of $\mathbf{a}$ were then switched back to be large (=$\mathbf{1}$, corresponding to a small recurrent gain) at the end of the trial, causing the output responses $\mathbf{y}$ to be extinguished. Target location was read out (at any time point during the delay period) by multiplying the responses with a pair of readout vectors:

$$\hat{\mathbf{x}} = \mathbf{W}_{ry}\mathbf{y} + \mathbf{c}_r = \mathbf{V}^t\mathbf{y} \tag{16}$$

$$\mathbf{V}^t = \begin{pmatrix} -0.5 & -0.3536 & 0 & 0.3536 & 0.5 & 0.3536 & 0 & -0.3536 \\ 0 & -0.3536 & 0.5 & 0.3536 & 0 & -0.3536 & -0.5 & -0.3536 \end{pmatrix},$$

where the rows of $\mathbf{W}_{ry} = \mathbf{V}^t$ were (same as the first two columns $\mathbf{W}_{zx}$) computed as the first two eigenvectors of the recurrent weight matrix $\mathbf{W}_{\hat{y}y}$., and $\mathbf{c}_r = \mathbf{0}$.

We also implemented a "batch algorithm" that optimizes **Eq. 1** for all time points at once. The batch algorithm is not intended to be a model of neural activity because it requires unlimited memory over time. But it is useful for testing and refining the optimization criterion (**Eq. 1**), and for debugging the incremental algorithm (**Eq. 13**) and the biophysical implementation (**Eqs. 57-71**). The batch algorithm works in two steps (analogous to back-propagation), a forward pass and a backward pass. The forward pass is expressed by the following system of discrete-time equations:

$$\begin{aligned}
\mathbf{z}(t) &= \mathbf{W}_{zx}\mathbf{x}(t) + \mathbf{c}_z \\
\hat{\mathbf{y}} &= \mathbf{W}_{\hat{y}y}\mathbf{y} + \mathbf{c}_{\hat{y}} \\
\Delta\boldsymbol{\alpha}(t) &= \frac{\Delta t}{\tau_\alpha}\Big[-\boldsymbol{\alpha}(t) + \mathbf{W}_{\alpha x}\mathbf{x}(t) + \mathbf{W}_{\alpha y}\mathbf{y}(t) + \mathbf{c}_\alpha\Big] \\
\boldsymbol{\alpha}(t+\Delta t) &= \boldsymbol{\alpha}(t) + \Delta\boldsymbol{\alpha}(t) \\
\Delta\mathbf{b}(t) &= \frac{\Delta t}{\tau_b}\Big[-\mathbf{b} + \mathbf{W}_{bx}\mathbf{x}(t) + \mathbf{W}_{by}\mathbf{y}(t) + \mathbf{c}_b\Big] \\
\mathbf{b}(t+\Delta t) &= \mathbf{b}(t) + \Delta\mathbf{b}(t)
\end{aligned} \tag{17}$$

The backward pass is:

$$\nabla E_j = \frac{\partial E}{\partial y_j(t)} = \left(\frac{b_j^*(t)}{1+b_j^*(t)}\right)\Big(y_j(t) - z_j(t)\Big) + \left(\frac{1}{1+b_j^*(t)}\right)\left(y_j(t) - \left(\frac{1}{1+\alpha_j^*(t)}\right)\hat{y}_j(t)\right) \tag{18}$$

$$\mathbf{y}(t) = \mathbf{y}(t) - r\vec{\nabla}E$$

The algorithm proceeds by alternating between the forward pass and the backward pass. For the batch algorithm, each of $\mathbf{x}$, $\mathbf{y}$, $\mathbf{z}$, $\mathbf{a}$, and $\mathbf{b}$ are stored as arrays (each is a vector for any given time point, over all time points), and the entire array (over all time points) is updated during each iteration. This is different from the incremental algorithm (**Eq. 13**) which stores only a vector of values for each of the variables ($\mathbf{x}$, $\mathbf{y}$, $\mathbf{z}$, $\mathbf{a}$, and $\mathbf{b}$), each of which is updated with each time step.

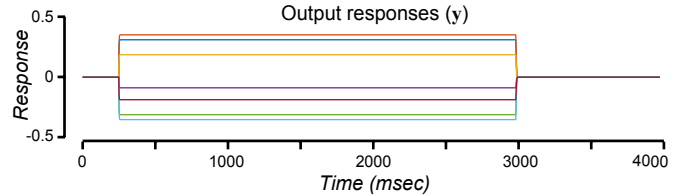

**Figure 3. Sustained delay-period activity, batch algorithm.** Output responses ($\mathbf{y}$) corresponding to the same inputs and weight matrices as in **Fig. 2**.

The dynamics of the responses (**Fig. 3**) are faster for the batch algorithm (compared to the incremental algorithm) because the batch algorithm does not include an intrinsic time constant $\tau_y$ for the output responses.

**Analysis: steady state, readout, and representational dimensionality**

The dynamics of the responses depend on the eigenvalues and eigenvectors of the recurrent weight matrix $\mathbf{W}_{\hat{y}y}$. The recurrent weight matrix depicted in **Fig. 2f** serves as a simple example. It is a symmetric, 8x8 matrix (N=8 is the number of neurons in that example). Two of the eigenvalues are equal to 1, two of them are 0, and the other 4 have values between 0 and 1. The weight matrix was in fact scaled so that the largest eigenvalues = 1. The corresponding eigenvectors define an orthogonal coordinate system (or basis) for the responses. The responses during the delay period (when $\mathbf{a}=\mathbf{0}$, $\mathbf{b}=\mathbf{0}$, $\mathbf{c}_z=\mathbf{0}$ and $\mathbf{c}_{\hat{y}}=\mathbf{0}$) are determined entirely by the projection of the initial values (the responses at the very beginning of the delay period) onto the eigenvectors. Eigenvectors with corresponding eigenvalues equal to 1 are sustained throughout the delay period. Those with eigenvalues less than 1 decay to zero (smaller eigenvalues decay more quickly). Those with eigenvalues greater than 1 would be unstable, growing without bound (which is why the weight matrix was scaled so that the largest eigenvalues = 1). So the steady-state responses during the delay period depend on the dot products of the initial responses and the two eigenvectors with corresponding eigenvalues equal to 1:

$$\mathbf{p} = \mathbf{V}^t\,\mathbf{y}_0 \qquad\qquad\qquad (19)$$

$$\mathbf{y}_s = \mathbf{V}\,\mathbf{p}\ ,$$

where $\mathbf{y}_s$ is the vector of steady-state responses, $\mathbf{y}_0$ is the vector of initial values at the beginning of the delay period, the rows of $\mathbf{V}^t$ were (**Eq. 16**) computed as the first two eigenvectors of the recurrent weight matrix $\mathbf{W}_{\hat{y}y}$, and $\mathbf{p}$ is the projection of $\mathbf{y}_0$ on $\mathbf{V}$. The same two eigenvectors were used to encode the input before the delay period:

$$\mathbf{y}_0 = \mathbf{V}\,\mathbf{x}_0\ , \qquad\qquad\qquad (20)$$

where the first two columns of $\mathbf{W}_{zx}$ are equal to $\mathbf{V}$, and $\mathbf{x}_0$ is a 2x1 vector corresponding to the target position. The same two eigenvectors were used to perform the readout (**Eq. 16**). Consequently, the readout recovers the input (substituting from **Eqs. 16** and **20** in **Eq. 19**)

$$\hat{\mathbf{x}} = \mathbf{V}^t\mathbf{y}_s = \mathbf{V}^t\mathbf{V}\mathbf{p} = \mathbf{V}^t\mathbf{V}\mathbf{V}^t\mathbf{y}_0 = \mathbf{V}^t\mathbf{V}\mathbf{V}^t\mathbf{V}\mathbf{x}_0 = \mathbf{x}_0\ \ , \qquad (21)$$

where the last step simplifies to $\mathbf{x}_0$ because $\mathbf{V}$ is an orthonormal matrix (i.e., $\mathbf{V}^t\mathbf{V} = \mathbf{I}$). The steady-state responses (and consequently the readout) are the same even when the encoding weights (the first two columns of $\mathbf{W}_{zx}$) also include components that are orthogonal to $\mathbf{V}$. Specifically, if the encoding weights are $\mathbf{V}+\mathbf{V}_p$ such that $\mathbf{V}^t\,\mathbf{V}_p = 0$:

$$\mathbf{y}_s = \mathbf{V}\mathbf{V}^t\mathbf{y}_0 = \mathbf{V}\mathbf{V}^t\left(\mathbf{V}+\mathbf{V}_p\right)\mathbf{x}_0 = \mathbf{V}\mathbf{V}^t\mathbf{V}\mathbf{x}_0 = \mathbf{V}\mathbf{x}_0$$

$$\hat{\mathbf{x}} = \mathbf{V}^t\mathbf{y}_s = \mathbf{V}^t\mathbf{V}\mathbf{x}_0 = \mathbf{x}_0 \qquad\qquad\qquad (22)$$

Likewise, the readout is unaffected by the offsets $\mathbf{c}_z$ and $\mathbf{c}_{\hat{y}}$, when they are orthogonal to $\mathbf{V}$.

The example ORGaNIC depicted in **Figs. 1** and **2** has a representational dimensionality $D = 2$, because the recurrent weight matrix has two eigenvalues equal to 1. This ORGaNIC is a two-dimensional continuous attractor during the delay period. It can maintain two values, corresponding to the horizontal and vertical locations of the target, where each of those values can be any real number.

**Manipulation via gated integration and reset**

Many behavioral tasks and AI applications require manipulation of information (i.e., working memory) in addition to maintenance of information (i.e., short-term memory). Such tasks and applications can take full advantage of the computational framework of ORGaNICs (by analogy with LSTMs), which dynamically change state depending on the current input and past context (via $\mathbf{a}$ and $\mathbf{b}$), in which the dependence on past inputs and outputs is controlled separately for each neuron (because the values of $\mathbf{a}$ and $\mathbf{b}$ can differ for each neuron). In addition, the encoding and readout weights may have components that are not orthogonal to $\mathbf{V}$, the offsets may have components that are not orthogonal to $\mathbf{V}$, and the inputs and context may change dynamically before the responses reach steady state. For example, if one of the components of $\mathbf{c}_{\hat{y}}$ is not orthogonal to $\mathbf{V}$, then the corresponding component of the responses will reflect the elapsed time interval since the beginning of the delay period.

A simulation of the double-step saccade task illustrates how ORGaNICs can both maintain and manipulate information over time (**Fig. 4**). In this task, two targets are shown while a subject is fixating the center of a screen (**Fig. 4a**, left panel). A pair of eye movements are then made in sequence to each of the two targets. Eye movements are represented in the brain using retinotopic, i.e., eye-centered coordinates (**Fig. 4a**, left panel, red lines). But after making the first eye movement, the plan for the second eye movement must be updated (**Fig. 4b**, middle panel, solid red line copied from left panel no longer points to the second target). This is done by combining a representation of the target location with a copy of the neural signals that control the eye muscles (called corollary discharge) to update the planned eye movement (**Fig. 4b**, middle panel, dashed red line). The ORGaNIC in **Fig. 4** received two types of inputs: 1) the target locations (**Fig. 4c,d**), and 2) the corollary discharge (**Fig. 4e**). The encoding weight matrix transformed the target positions to the responses of a network of 16 neurons (8 for each target). As in **Fig. 2**, the responses $\mathbf{y}$ followed the input drive $\mathbf{z}$ at the beginning of the simulated trial because $\mathbf{a}$ and $\mathbf{b}$ were large (=$\mathbf{1}$, corresponding to a short effective time constant). The values of $\mathbf{a}$ and $\mathbf{b}$ were then switched to be small (=$\mathbf{0}$, corresponding to a long effective time constant) before the targets were extinguished, so the output responses $\mathbf{y}$ exhibited sustained delay-period activity that represented the original target locations (**Fig. 4c-k**, leftmost vertical dashed line). The values of $\mathbf{a}$ were then switched back to be large (=$\mathbf{1}$, corresponding to a small recurrent gain) at the end of the trial, causing the output responses $\mathbf{y}$ to be extinguished. The modulators $\mathbf{b}$ were non-zero during eye movements (**Fig. 4g**), when the corollary discharge was non-zero (**Fig. 4c**), so that the responses $\mathbf{y}$ integrated the corollary discharge with the stored representation of the target locations. The last two panels (**Fig. 4j,k**) show the readout of each of the two target locations over time. Preceding the first eye movement (**Fig. 4j,k**, leftmost dashed line), the responses encoded the original target locations: $(1, 0.5)$ and $(-1, 0.5)$. After the first eye movement (**Fig. 4j,k**, middle dashed line), the responses encoded the corresponding updated target locations: $(0, 0)$ and $(0, -1)$. After the second eye movement (**Fig. 4j,k**, rightmost dashed line), the responses were updated again to encode correspondingly updated target locations: $(0, 1)$ and $(0, 0)$. The corollary discharge (**Fig. 4e**) was simply a copy of the readout from just before the corresponding eye movement, i.e., from **Fig. 4j** just before the first eye movement and from **Fig. 4k** just before the second eye movement.

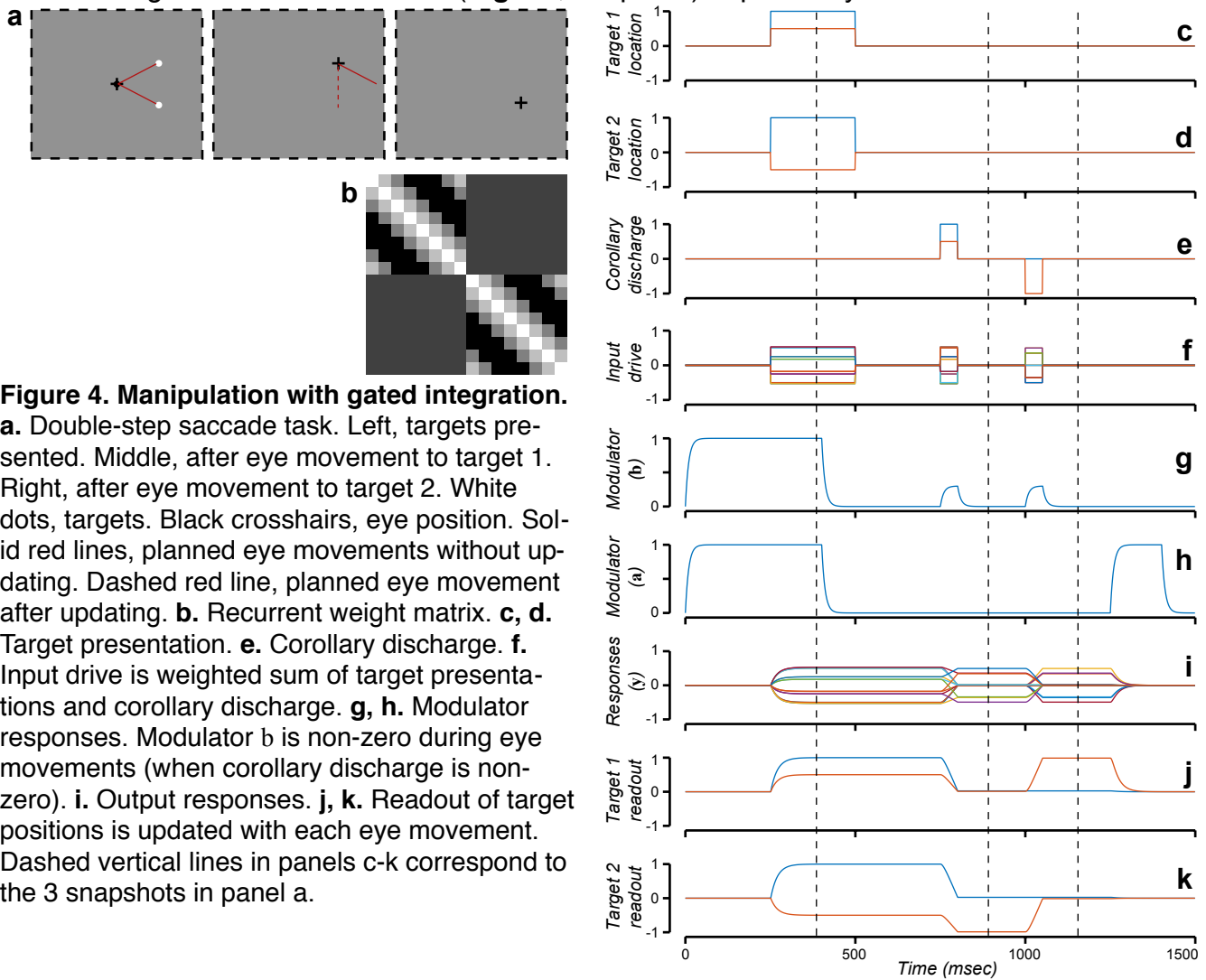

**Figure 4. Manipulation with gated integration.**
**a.** Double-step saccade task. Left, targets presented. Middle, after eye movement to target 1. Right, after eye movement to target 2. White dots, targets. Black crosshairs, eye position. Solid red lines, planned eye movements without updating. Dashed red line, planned eye movement after updating. **b.** Recurrent weight matrix. **c, d.** Target presentation. **e.** Corollary discharge. **f.** Input drive is weighted sum of target presentations and corollary discharge. **g, h.** Modulator responses. Modulator $\mathbf{b}$ is non-zero during eye movements (when corollary discharge is non-zero). **i.** Output responses. **j, k.** Readout of target positions is updated with each eye movement. Dashed vertical lines in panels c-k correspond to the 3 snapshots in panel a.

## Oscillatory activity and complex dynamics

The same computational framework can be used to generate delay-period activity with complex dynamics, rather than just sustained activity, and the same theoretical framework can be used to analyze it. The key idea is that the weights and the output responses may be complex-valued. The complex-number notation is just a notational convenience. The complex-valued responses may be represented by pairs of neurons, and the complex-valued weights in the recurrent weight matrix may be represented by pairs of synaptic weights.

A recurrent weight matrix in the form of a synfire chain (Abeles 1991; Abeles 1982; Abeles et al. 1993; Bienenstock 1995; Herrmann et al. 1995), for example, generates periodic, sequential activity (**Fig. 5**). The example network depicted in **Fig. 5** has 100 neurons, the recurrent weight matrix $\mathbf{W}_{\hat{y}y}$ (a 100x100 matrix) contains 0's along the main diagonal and 1's along the diagonal adjacent to the main diagonal (**Fig. 5a**), looping back from the 100th neuron to the 1st neuron (**Fig. 5a**, top-right). Consequently, the activity is "handed off" from one neuron to the next during a delay period. The encoding matrix is 100x4. Otherwise, this example network is the same as that depicted in **Figs 1** and **2**. The responses are complex-valued and oscillatory during the delay period (**Fig. 5b**).

The dynamics of the responses during the delay period again depend on the eigenvalues and eigenvectors of the recurrent weight matrix $\mathbf{W}_{\hat{y}y}$. For this recurrent weight matrix, the eigenvectors and eigenvalues are complex-valued. Two of the eigenvalues have real parts equal to 1. The rest of the eigenvalues have real parts that are less than 1 (some of them being negative). The components of the responses corresponding to the 98 eigenvalues with real parts that are less than 1 decay to 0 during a delay period, so only the first two components are relevant for the steady-state responses. The imaginary parts of the first two eigenvalues determine the oscillation frequencies. In this example, the imaginary parts are small (±0.0629), corresponding to a slow frequency. In fact, the period of oscillation is ~1 sec because there are 100 neurons in the synfire chain, each with an intrinsic time-constant $\tau_y = 10$ ms. In spite of the oscillations, target location may be read out (up to a sign change), at any time point during the delay period, by multiplying the responses with a pair of readout vectors:

$$\hat{\mathbf{x}} = \left| \mathbf{V}^t \mathbf{y} \right| = \left| \mathbf{x}_0 \right| \ , \tag{23}$$

where the superscript $t$ now means conjugate transpose, the rows of $\mathbf{V}^t$ are once again the same as the first two columns of the encoding matrix $\mathbf{W}_{zx}$, computed as the first two (complex-valued) eigenvectors of the recurrent weight matrix $\mathbf{W}_{\hat{y}y}$. Only the absolute value of the input can be read out (i.e., up to a sign change) using **Eq. 23**, because the responses oscillate over time with a combination of frequency components (depending on the imaginary parts of the eigenvectors), and the frequencies, phases, and elapsed time are presumed to be unknown. The sign may be resolved if the frequencies, phases, and elapsed time are known.

**Figure 5. Periodic dynamics with synfire chain recurrent weights. a.** Recurrent weight matrix ($\mathbf{W}_{\hat{y}y}$). N=100; d=2. **b.** Output responses (**y**). Upper panel, real part of the complex-valued responses. Lower panel, imaginary part. Only a small fraction of the 100 response time-courses are shown in each panel.

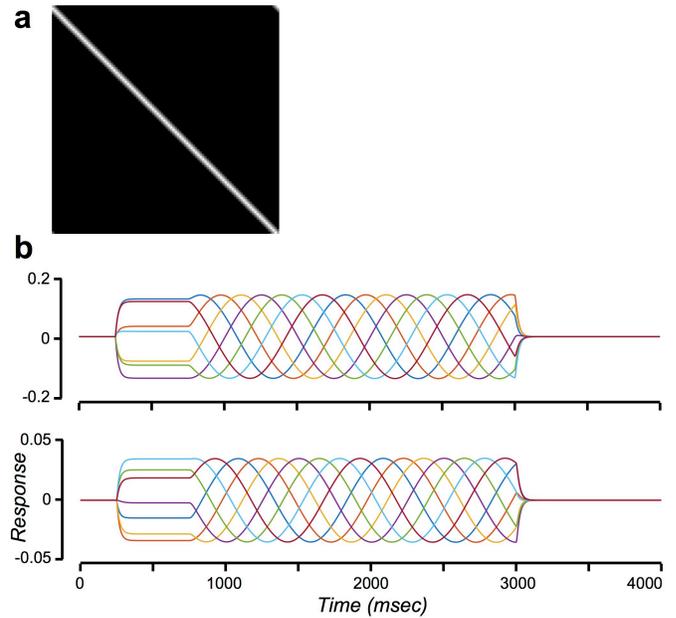

A further generalization is depicted in **Fig. 6**, which exhibits complex dynamics. Once again, in spite of the complex dynamics, target location may be read out (up to a sign change), at any point during the delay period (**Eq. 23**). In this example, there are again 100 neurons and the recurrent weight matrix $\mathbf{W}_{\hat{y}y}$ is a 100x100 matrix. The recurrent weight matrix was designed to have real parts of 10 eigenvalues equal to 1, real parts of the other 90 eigenvalues less than 1, and with small imaginary parts (between -0.1 and 0.1) for all 100 eigenvalues. Consequently, the steady-state responses have a representational dimensionality of $D = 10$, $\mathbf{V}$ is a 100x10 matrix containing 10 eigenvectors, and $\mathbf{W}_{zx}$ is a 100x12 matrix (10 dimensions plus 2 cues).

The recurrent weight matrix was created using the following algorithm. First, we filled a 100x100 matrix $\mathbf{A}$ with random complex-valued numbers (normally distributed with mean 0 and standard deviation 1). Second, we computed the QR decomposition of the matrix $\mathbf{A}$, yielding a unitary matrix of complex-valued numbers $\mathbf{Q}$, such that $\mathbf{Q}^t \, \mathbf{Q} = \mathbf{Q} \, \mathbf{Q}^t = \mathbf{I}$. Third, we created a diagonal matrix $\mathbf{D}$ with the desired eigenvalues, with real parts equal to 1 for 10 of them and random numbers (uniformly distributed between 0 and 1) for the rest, and with random imaginary parts (normally distributed with mean 0 and standard deviation 0.05). Again, the imaginary parts of the eigenvalues were purposefully chosen to be small so that the oscillation frequencies would be low. Finally, we multiplied $\mathbf{W}_{\hat{y}y} = \mathbf{Q} \, \mathbf{D} \, \mathbf{Q}^t$, similar to reconstructing a matrix from its singular-value decomposition.

**Figure 6. Delay-period dynamics with general recurrent weights. a.** Recurrent weight matrix ($\mathbf{W}_{\hat{y}y}$). Left, real part of the complex-valued weight matrix (values of the weights range from -0.1 to 0.7; white, positive weights; black, negative weights). Right, imaginary part (values of the weights range from -0.1 to 0.1; white, positive weights; black, negative weights). **b.** Output responses ($\mathbf{y}$). Upper panel, real part of the complex-valued responses. Lower panel, imaginary part. Only a small fraction of the 100 response time-courses are shown in each panel.

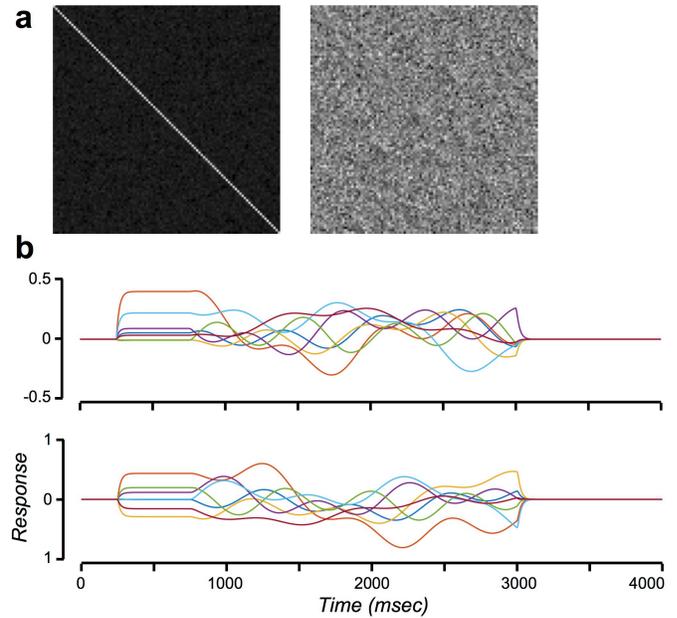

This example ORGaNIC has a representational dimensionality $D = 10$, because the recurrent weight matrix was constructed to have 10 eigenvalues with real parts equal to 1. It is a ten-dimensional continuous attractor during the delay period, and it can maintain ten values (e.g., the horizontal and vertical locations of 5 targets). In the preceding examples (**Figs. 2** and **4**, with center-surround recurrent weights or synfire weights, respectively), the recurrent weight matrices have only 2 eigenvalues with real parts equal to 1, so the representational dimensionality of those ORGaNICs is only $D = 2$.

The neurons in each example ORGaNIC, thus far, all have the same time constant. But that can be generalized, e.g., so that inhibitory neurons have a slower time constant (and consequently delayed responses) compared to the excitatory neurons:

$$\tau_{y_j} \frac{dy_j}{dt} = -y_j + \left(\frac{b_j^+}{1+b_j^+}\right) z_j + \left(\frac{1}{1+a_j^+}\right) \hat{y}_j \; , \tag{24}$$

where the subscript $j$ on where $\tau_y$ specifies the intrinsic time-constant for each neuron. The responses during the delay period (when $\mathbf{a}{=}\mathbf{0}$, $\mathbf{b}{=}\mathbf{0}$, $\mathbf{c}_z{=}\mathbf{0}$ and $\mathbf{c}_{\hat{y}}{=}\mathbf{0}$) are:

$$d\left(\boldsymbol{\tau}_y\right) \frac{d\mathbf{y}}{dt} = -\mathbf{y} + \mathbf{W}_{\hat{y}y} \mathbf{y} \; , \tag{25}$$

where $d(\boldsymbol{\tau})$ takes a vector $\boldsymbol{\tau}$ and turns it into a diagonal matrix. The eigenvalues of the recurrent weight matrix, combined with the value(s) of the time constant(s) determine whether or not there is sustained activity or oscillations, whether the oscillations are stable or decaying, and the frequencies of the oscillations. Simplifying **Eq. 25**:

$$d\left(\boldsymbol{\tau}_y\right)\frac{d\mathbf{y}}{dt} = -\mathbf{y} + \mathbf{W}_{\hat{y}y}\mathbf{y}$$

$$d\left(\boldsymbol{\tau}_y\right)\frac{d\mathbf{y}}{dt} = \left(\mathbf{W}_{\hat{y}y} - \mathbf{I}\right)\mathbf{y}$$

$$\frac{d\mathbf{y}}{dt} = d\left(\tfrac{1}{\boldsymbol{\tau}_y}\right)\left(\mathbf{W}_{\hat{y}y} - \mathbf{I}\right)\mathbf{y}$$

$$\frac{d\mathbf{y}}{dt} = \mathbf{W}'_{\hat{y}y}\mathbf{y}$$

$$,\qquad(26)$$

where $\mathbf{I}$ is the identity matrix. The matrix $\mathbf{W}'_{\hat{y}y}$ depends on both the recurrent weight matrix and the time constants:

$$\mathbf{W}'_{\hat{y}y} = d\left(\tfrac{1}{\boldsymbol{\tau}_y}\right)\left(\mathbf{W}_{\hat{y}y} - \mathbf{I}\right)\ . \qquad(27)$$

If an eigenvalue of $\mathbf{W}'_{\hat{y}y}$ has real part equal to zero and non-zero imaginary part then the responses exhibit stable, ongoing oscillations. The frequency of the oscillations is specified by the imaginary part of the eigenvalue:

$$f_i = \tfrac{1000}{2\pi}\,\mathrm{Im}\left(\lambda_i\right) \qquad(28)$$

where $\lambda_i$ is the imaginary part of the $i^{\text{th}}$ eigenvalue of $\mathbf{W}'_{\hat{y}y}$, $f_i$ is the corresponding oscillation frequency (in Hz), and the factor of 1000 is needed because the time constants $\tau_{yj}$ are presumed to be specified in msec. If the time constants are all equal to one another (i.e., **Eq. 24** reduces to **Eq. 9**), then the real part of an eigenvalue of $\mathbf{W}_{\hat{y}y}$ equals 1 when the corresponding eigenvalue of $\mathbf{W}'_{\hat{y}y}$ equals 0. But there are some cases in which the real part of an eigenvalue of $\mathbf{W}'_{\hat{y}y}$ equals 0 even when the corresponding eigenvalue of $\mathbf{W}_{\hat{y}y}$ is less 1.

This analysis of the recurrent weight matrix and time constants, can be applied to generalize the notion of E:I balance in a cortical circuit (**Fig. 7**). For example, consider the following recurrent weight matrix:

$$\mathbf{W}_{\hat{y}y} = \begin{pmatrix} 2 & -1 \\ 2 & -0.25 \end{pmatrix}, \qquad(29)$$

which corresponds to an E:I pair of neurons; the first neuron is excitatory (both weights in the first column are positive) and the second neuron is inhibitory (both weights in the second column are negative). The largest eigenvalue of this recurrent weight matrix has a real part that is less than one and a non-zero imaginary part. If the two time constants are equal to one another, then the responses are damped oscillators that decay over time during a delay period (**Fig. 7a**). If the time constant of the inhibitory neuron is 25% longer than the excitatory neuron, then the real part of the largest eigenvalue of $\mathbf{W}'_{\hat{y}y}$ equals equals 0 (even though the largest eigenvalue of $\mathbf{W}_{\hat{y}y}$ is less than 1). When this is the case, the responses exhibit stable, ongoing oscillations (**Fig. 7b**). Scaling the time constants $\tau_{yj}$ of the neurons, all by the same factor, scales the oscillation frequencies by the inverse of that scale factor (**Fig. 7c**).

**Figure 7. Stability depends and recurrent weights and time constants. a.** Responses of E:I pair of neurons with recurrent weight matrix given by **Eq. 29**. Time courses of input and modulators was the same as in **Fig. 2**. Blue, excitatory neuron. Orange, Inhibitory neuron. Time constants: 10 ms and 10 ms. **b.** Responses exhibit stable, ongoing oscillations with time delay between excitation and inhibition. Time constants: 10 ms and 12.5 ms. Oscillation frequency = 12 Hz. **c.** Oscillation frequency is decreased by scaling the time constants. Time constants: 20 ms and 25 ms. Oscillation frequency = 6 Hz.

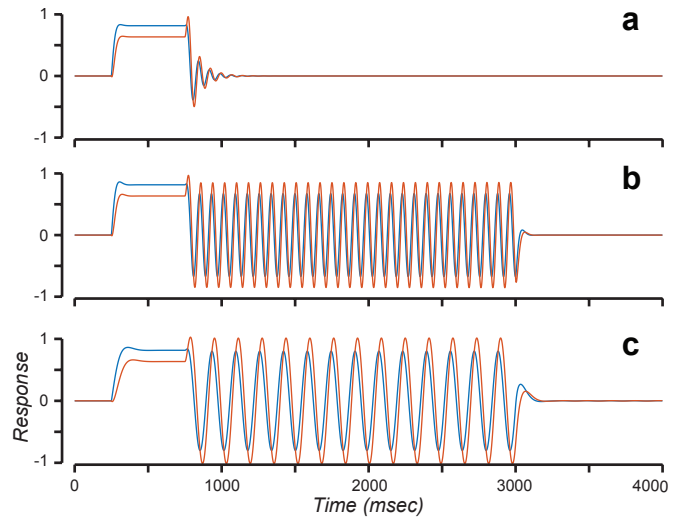

**Signal generators, motor preparation, and motor control**

ORGaNICs are also capable of generating signals, like that needed to execute a complex sequence of movements (e.g., bird song, speech generation, golf swing, bowling, skiing moguls, backside double McTwist 1260 on a snowboard out of the halfpipe). Some actions are ballistic (open loop), meaning that they are executed with no sensory feedback. Others are closed loop, meaning that the movements are adjusted on the fly based on sensory feedback. ORGaNICs can produce patterns of activity over time that might underlie the execution of both open- and closed-loop movements.

Open-loop control corresponds to the delay-period responses described above. Each eigenvector of the recurrent weight matrix is associated with a basis function, a pattern of activity across the population of neurons and over time. The basis functions are complex exponentials (sines and cosines) when the modulators ($\mathbf{a}$ and $\mathbf{b}$) are 0. Examples of these basis functions are plotted in **Fig. 8a,b**. The time course of the input, and the time courses of the modulators ($\mathbf{a}$ and $\mathbf{b}$) were the same in this simulation as in **Figs. 2, 3,** and **5**, and the recurrent weight matrix was the same as that shown in **Fig. 6a**. When the input drives only a single eigenvector (i.e., because the input is orthogonal to the other eigenvectors), then the respons-es during the delay period (when $\mathbf{a} = \mathbf{b} = \mathbf{0}$) are sinusoidal, all with the same frequency (as determined by the imaginary part of the corresponding eigenvalue). But the response of each neuron, in general, exhibits a different amplitude and phase (**Fig. 8a**). When the input drives another eigenvector, then the neurons exhibit responses with a different temporal frequency (**Fig. 8b**). A linear sum of the inputs evokes a linear sum of the responses (compare **Fig. 8c** and **Fig. 8d**). Linear sums of these sinusoidal basis functions may be used as control signals for ballistic (open loop) movements. The basis functions are damped oscillators when the modulators are greater than 0 but equal to one another ($\mathbf{a} = \mathbf{b}$) and constant over time, and when the input is constant over time.

**Figure 8. Signal generators for motor preparation and motor control. a.** Response time-series when input (with amplitude = 1) drives only one of the eigenvectors of the recurrent weight matrix ($\mathbf{W}_{\hat{y}y}$). Left panel, real part of the complex valued responses ($\mathbf{y}$). Right panel, imaginary part of the complex-valued responses. **b.** Response time-series when input drives a second eigenvector. Same format as panel a. **c.** Response time course driven by a linear combination of the inputs in panels a and b. **d.** Linear combination of the responses from panels a and b. The curves in panel d are identical those in panel c because the responses behave like a linear system; a linear sum of the inputs evokes a linear sum of the responses. Only a small fraction of the 100 response time-courses are shown in each panel.

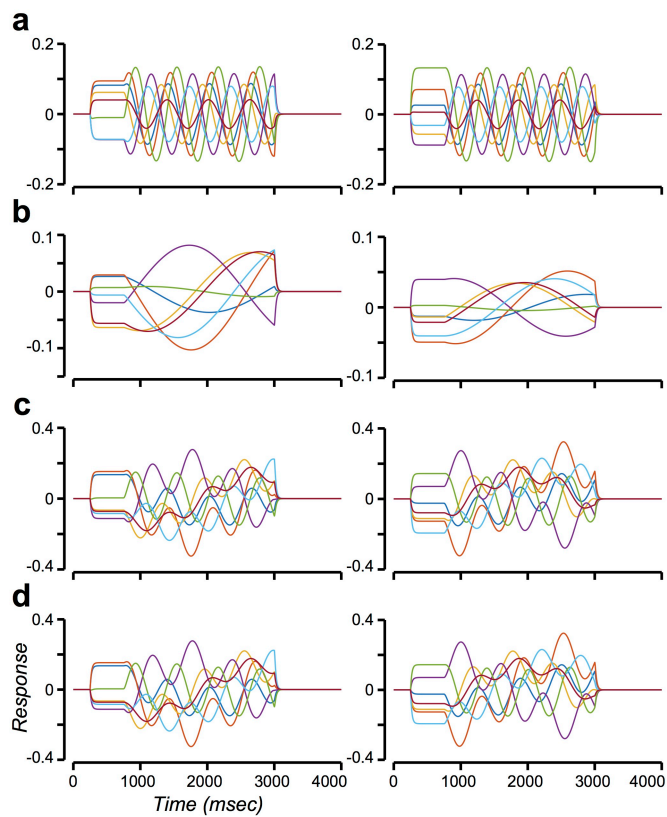

The readout for open-loop control is, in general, an arbitrary linear transform of the responses: $\mathbf{W}_{ry}\, \mathbf{y}$ (optionally followed by an output nonlinearity). The readout matrix for short-term memory, as discussed above, may be comprised of the eigenvectors of the recurrent weight matrix. Doing so ensures that the input is recovered (up to a sign change) at any time during the delay period. But recovering the input is not the goal for open-loop control.

If the input is varying over time, then the responses depend on a linear combination of the inputs and the basis functions, and the responses may be used for closed-loop control. If the modulators ($\mathbf{a}$ and $\mathbf{b}$) are also time-varying, then the closed-loop control responses may exhibit a wide range of complex dynamics.

**Biophysical implementation**

A possible biophysical implementation of ORGaNICs is depicted in **Fig. 9**. The key idea is that the two terms corresponding to the input drive and recurrent drive are computed in separate dendritic compartments of a cortical pyramidal cell. First, we hypothesize that the input drive $\mathbf{z}$, modulated by $\mathbf{b}$, is computed in the soma and basal dendrites. Second, we hypothesize that the recurrent drive $\hat{\mathbf{y}}$, modulated by $\mathbf{a}$, is computed in the apical dendrites. Furthermore, we assert, based on experimental evidence (Fuster and Alexander 1971; Guo et al. 2017; Schmitt et al. 2017), that the modulatory responses corresponding to the modulators, $\mathbf{a}$ and $\mathbf{b}$, are computed in the thalamus.

The analysis relies on an equivalent circuit model of a pyramidal cell. Heeger's lecture notes provide an introduction to how we approach neural computation with equivalent circuit models:

http://www.cns.nyu.edu/~david/handouts/membrane.pdf

http://www.cns.nyu.edu/~david/handouts/synapse.pdf

A simplified electrical-circuit model of a pyramidal cell is depicted in **Fig. 9a**. The model comprises 3 compartments for the soma, the apical dendrite, and the basal dendrite. Each compartment is an RC circuit with a variable-conductance resistor and a variable current source. The capacitors represent the electrical capacitance of the neural membrane. Each current source approximates a combination of excitatory and inhibitory synaptic inputs (see above link to lecture notes). Each variable-conductance resistor ($g_{va}$, $g_{vb}$, and $g_{vs}$) represents shunting synaptic input. The two fixed-conductance resistors ($R_a$ and $R_b$) represent the resistances between the compartments (i.e., along the dendritic shafts).

*Pyramidal cell model*

To analyze the function of this electrical-circuit model, we express the somatic membrane potential as a function of the synaptic inputs (the variable conductance resistors and the cur-rent sources in each compartment). To simplify the derivation, without loss of generality, we set the resting membrane potential (corresponding to when there is no synaptic input) to 0. (For a non-zero resting potential, we could simply add a voltage source, i.e., battery, equal to the resting potential, in series with variable-conductance resistors in each compartment.) From Kirchhoff's current law:

$$C\frac{dv_a}{dt} + g_{va}v_a - I_a + I_{as} = 0 \tag{30}$$

$$C\frac{dv_b}{dt} + g_{vb}v_b - I_b + I_{bs} = 0 \tag{31}$$

$$C\frac{dv_s}{dt} + g_{vs}v_s - I_s - I_{as} - I_{bs} = 0 \quad . \tag{32}$$

**Figure 9. Biophysical implementation. a.** Electrical-circuit model of a pyramidal cell with separate RC circuit compartments for the soma, the apical dendritic tree, and the basal dendritic tree. $g_{va}$, $g_{vb}$, shunting synapses represented by variable-conductance resistors. $I_a$, $I_b$, synaptic input represented by variable current sources. $v_s$, somatic membrane potential. $v_a$, $v_b$, membrane potentials, respectively, in apical and basal dendrites. $C$, membrane capacitance. **b.** Electrical-circuit models of thalamic cells corresponding to the modulators. $a$, $b$, membrane potentials of two neurons corresponding to the two modulators. $g_{ea}$, $g_{eb}$, excitatory synaptic conductances. $g_{ia}$, $g_{ib}$, inhibitory synaptic conductances. $E_e$, reversal potential of excitatory synapses. $E_i$, reversal potential of inhibitory synapses. $g_l$, leak conductance. **c.** Output responses ($\mathbf{y}$) corresponding to the same inputs and weight matrices as in **Fig. 2**. Model parameters: $C = 1$; $g_{vs} = 1$; $R_a = 10$; $R_b = 1$; $g_l = 1$.

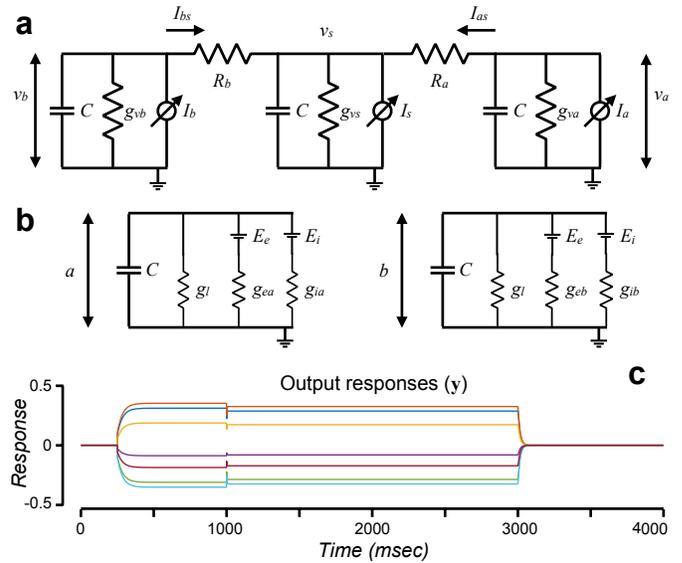

The membrane potentials $v_s$, $v_a$, and $v_b$ correspond to the soma, the apical dendrite, and the basal dendrite, respectively. The currents flowing from the soma, and from each of the dendrites, to the extracellular space are denoted $I_s$, $I_a$, and $I_b$. The internal currents flowing between compartments are denoted $I_{as}$ and $I_{bs}$. In addition, from Ohm's Law:

$$v_a - v_s = R_a I_{as} \tag{33}$$

$$v_b - v_s = R_b I_{bs} \quad . \tag{34}$$

Substituting for $I_{as}$ and $I_{bs}$. in **Eq. 32** from **Eqs. 30** and **31**:

$$\left( C\frac{dv_s}{dt} + g_{vs}v_s - I_s \right) + \left( C\frac{dv_a}{dt} + g_{va}v_a - I_a \right) + \left( C\frac{dv_b}{dt} + g_{vb}v_b - I_b \right) = 0 \quad . \tag{35}$$

The steady-state values for the membrane potentials and internal currents, assuming that the synaptic inputs are constant over time, are derived by setting the derivatives equal to zero in

**Eqs. 30**, **31**, and **35**:

$$-g_{va}v_a + I_a = I_{as} \tag{36}$$

$$-g_{vb}v_b + I_b = I_{bs} \tag{37}$$

$$g_{vs}v_s - I_s + g_{va}v_a - I_a + g_{vb}v_b - I_b = 0 \quad . \tag{38}$$

Substituting for the internal current $I_{as}$ from **Eq. 36** into **Eq. 33**:

$$
\begin{aligned}
v_a - v_s &= R_a I_{as} \\
v_a - v_s &= R_a \left( I_a - g_{va}v_a \right) \\
v_a + R_a g_{va} v_a &= v_s + R_a I_a \\
(1 + R_a g_{va})v_a &= v_s + R_a I_a \\
v_a &= \frac{v_s + R_a I_a}{1 + R_a g_{va}} = \frac{v_s}{1 + R_a g_{va}} + \frac{R_a I_a}{1 + R_a g_{va}}
\end{aligned}
\tag{39}
$$

Likewise, substituting for the internal current $I_{bs}$ from **Eq. 37** into **Eq. 34**:

$$v_b = \frac{v_s}{1 + R_b g_{vb}} + \frac{R_b I_b}{1 + R_b g_{vb}} \quad . \tag{40}$$

Substituting for $v_a$ and $v_b$. from **Eqs. 39-40** into **Eq. 38**:

$$
\begin{aligned}
g_{vs}v_s - I_s + \frac{g_{va}}{1 + R_a g_{va}}v_s + \frac{R_a g_{va}}{1 + R_a g_{va}}I_a - I_a + \frac{g_{vb}}{1 + R_b g_{vb}}v_s + \frac{R_b g_{vb}}{1 + R_b g_{vb}}I_b - I_b &= 0 \\
v_s \left( g_{vs} + \frac{g_{va}}{1 + R_a g_{va}} + \frac{g_{vb}}{1 + R_b g_{vb}} \right) = I_s + I_a \left( 1 - \frac{R_a g_{va}}{1 + R_a g_{va}} \right) + I_b \left( 1 - \frac{R_b g_{vb}}{1 + R_b g_{vb}} \right) \\
v_s \left( g_{vs} + \frac{g_{va}}{1 + R_a g_{va}} + \frac{g_{vb}}{1 + R_b g_{vb}} \right) = I_s + \left( \frac{1}{1 + R_a g_{va}} \right) I_a + \left( \frac{1}{1 + R_b g_{vb}} \right) I_b
\end{aligned}
\tag{41}
$$

**Eq. 41** is an expression for the steady-state somatic membrane potential $v_s$ in terms of the synaptic inputs ($I_s$, $I_a$, $I_b$, $g_{va}$, $g_{vb}$, and $g_{vs}$) and the fixed (constant) resistances along the dendrites ($R_a$ and $R_b$).

To implement ORGaNICs with this pyramidal-cell model, we specify the synaptic inputs ($I_s$, $I_a$, $I_b$, $g_{va}$, $g_{vb}$, and $g_{vs}$) to each neuron in terms of its input drive ($y$), recurrent drive ($\hat{y}$), and modulators ($a$ and $b$):

$$
\begin{aligned}
g_{va}(t) &= \tfrac{1}{R_a} a^+(t) \\
g_{vb}(t) &= \tfrac{1}{R_b} b^+(t) \\
I_s(t) &= z(t) \\
I_b(t) &= -z(t) \\
I_a(t) &= \hat{y}(t) = y^+(t) - y^-(t)
\end{aligned}
\tag{42}
$$

where $g_{vs}$ is presumed to be a constant. We presume that the output firing rate is well-approximated by halfwave rectification:

$$y^+(t) = \lfloor v_s(t) \rfloor \quad , \tag{43}$$

and that the negative values (corresponding to hyperpolarlization of the membrane potential $v_s$) are represented by a separate neuron that receives the complementary synaptic inputs (identical for $g_{va}$ and $g_{vb}$, and opposite in sign for $I_s$, $I_a$, and $I_b$), analogous to ON- and OFF-center retinal ganglion cells. Substituting from **Eq. 42** into **Eq. 41**:

$$v_s \left( g_{vs} + \frac{a^+}{R_a\left(1+a^+\right)} + \frac{b^+}{R_b\left(1+b^+\right)} \right) = z + \frac{1}{1+a^+}\hat{y} - \frac{1}{1+b^+}z \quad , \tag{44}$$

$$g_v v_s = \frac{b^+}{1+b^+}z + \frac{1}{1+a^+}\hat{y} \quad , \tag{45}$$

where $g_v$ is the total synaptic conductance:

$$g_v = g_{vs} + \frac{a^+}{R_a\left(1+a^+\right)} + \frac{b^+}{R_b\left(1+b^+\right)} \quad . \tag{46}$$

The steady-state membrane potential (**Eq. 45**) is a weighted sum of the input drive and recurrent drive, modulated by $a$ and $b$, respectively, and then scaled by the total somatic conductance. This is identical to the steady-state response of an ORGaNIC (compare **Eq. 45** with **Eq. 9**) when the total somatic conductance is $g_v = 1$.

There are a variety of combinations of the various parameters of this model for which the total somatic conductance is approximately equal to 1. Two particular interesting special cases correspond to when the modulators are both on (e.g., when the responses are dominated by the input drive), and when the modulators are both off (e.g., delay period). The first special case is:

For $g_{vs} = 1$, $a^+ \ll 1$, $b^+ \ll 1$, $R_a \geq 1$, $R_b \geq 1$:

$$g_v \approx 1 + \frac{a^+}{R_a} + \frac{b^+}{R_b} = \frac{R_a R_b + R_b a^+ + R_a b^+}{R_a R_b} \quad .$$

$$\frac{1}{g_v} \approx \frac{R_a R_b}{R_a R_b + R_b a^+ + R_a b^+} \approx 1 \tag{47}$$

The second special case is:

For $g_{vs} = 1$, $a^+ \gg 1$, $b^+ \gg 1$, $R_a R_b \gg 1$:

$$g_v \approx 1 + \frac{1}{R_a} + \frac{1}{R_b} = \frac{R_a R_b + 2}{R_a R_b} \quad .$$

$$\frac{1}{g_v} \approx \frac{R_a R_b}{R_a R_b + 2} \approx 1 \tag{48}$$

This is, of course, a simplified model of pyramidal cells. There are two simplifications that are particularly suspect. First is that there is no leak conductance in the dendrites. We can think of $g_{vs} = 1$ as the somatic leak conductance, but the shunt conductances in the dendrites must be allowed to go to zero to correctly implement ORGaNICs. The second approximation is that the input drive and recurrent drive are mediated by synaptic currents, not conductance changes. A push-pull arrangement of synaptic inputs can act like a current source (Carandini and Heeger 1994). See also: http://www.cns.nyu.edu/~david/handouts/membrane.pdf. But doing so necessitates a high level of spontaneous activity so that increases in excitation are met with equal decreases in inhibition, and vice versa, whereas spontaneous activity in PFC (and parietal cortex) is generally low. Synaptic inputs can also be approximated as current sources when the membrane potential remains far from the (excitatory and inhibitory) synaptic reversal potentials.

See Appendix for full details of the implementation.

**Prediction**

Information processing in the brain is dynamic; dynamic and predictive processing is needed to control behavior in sync with or in advance of changes in the environment. Without prediction, behavioral responses to environmental events will always be too late because of the lag or latency in sensory and motor processing. Prediction is a key component of theories of motor control and in explanations of how an organism discounts sensory input caused by its own behavior (e.g., Crapse and Sommer 2008; Kawato 1999; Wolpert et al. 1995). Prediction has also been hypothesized to be essential in sensory and perceptual processing (e.g., Bialek et al. 2001; Nijhawan 2008; Palmer et al. 2015).

A variant of the ORGaNICs computational motif is capable of prediction over time, following our previous work (Heeger 2017). Once again, the starting point is an optimization criterion (or energy function) that represents a compromise between between the input drive and the recurrent drive, over time:

$$
E = \frac{1}{2} \int_t \sum_j \left( \frac{b_j^+}{1+b_j^+} \right) \left[ \sum_k \mathrm{Re}\left(y_k\right) - x \right]^2 + \left( \frac{1}{1+b_j^+} \right) \left[ y_j - \left( \frac{1}{1+\alpha_j^+} \right) \hat{y}_j \right]^2
$$

$$
\propto \frac{1}{2} \sum_t \sum_j \left( \frac{b_j^+}{1+b_j^+} \right) \left[ \sum_k \mathrm{Re}\left(y_k\right) - x \right]^2 + \left( \frac{1}{1+b_j^+} \right) \left[ y_j - \left( \frac{1}{1+\alpha_j^+} \right) \hat{y}_j \right]^2
$$

(49)

$$
\alpha_j^+ \geq 0 \ \text{ and } \ b_j^+ \geq 0 \ ,
$$

where the superscript + is a rectifying output nonlinearity. The second term in this equation is the same as **Eq. 1**. The first term in **Eq. 49** constrains the sum of the output responses to be similar to the input $x$. As expressed here, the input is presumed to be real-valued which is why the real parts of the output responses are summed, but it would of course be trivial to sum the complex-valued output responses so as to handle complex-valued inputs. The neural responses are again (analogous to **Eq. 2**) modeled as dynamical processes that minimize the energy $E$ over time:

$$
\tau_y \frac{dy_j}{dt} = -\frac{dE}{dy_j}
$$

(50)

$$
= -\left( \frac{b_j^+}{1+b_j^+} \right) \left[ \sum_k \mathrm{Re}\left(y_k\right) - x \right] - \left( \frac{1}{1+b_j^+} \right) \left[ y_j - \left( \frac{1}{1+\alpha_j^+} \right) \hat{y}_j \right]
$$

$$
= -y_j + \left( \frac{b_j^+}{1+b_j^+} \right) x + \left( \frac{1}{1+b_j^+} \right) \left( \frac{1}{1+\alpha_j^+} \right) \hat{y}_j + \left( \frac{b_j^+}{1+b_j^+} \right) \left[ y_j - \sum_k \mathrm{Re}\left(y_k\right) \right] .
$$

Following **Eq. 9**, we again introduce a change of variables:

$$
\tau_y \frac{dy_j}{dt} = -y_j + \left( \frac{b_j^+}{1+b_j^+} \right) x + \left( \frac{1}{1+\alpha_j^+} \right) \hat{y}_j + \left( \frac{b_j^+}{1+b_j^+} \right) \left[ y_j - \sum_k \mathrm{Re}\left(y_k\right) \right] .
$$

(51)

If the input $x$ is complex-valued, as noted above, then the last term depends on the sum of the complex-valued responses, not just the real parts, yielding:

$$\tau_y \frac{dy_j}{dt} = -y_j + \left(\frac{b_j^*}{1+b_j^*}\right)x + \left(\frac{1}{1+a_j^*}\right)\hat{y}_j - \left(\frac{b_j^*}{1+b_j^*}\right)\left(\sum_{k \neq j} y_k\right) . \tag{52}$$

The first and third terms in each of **Eqs. 51** and **52** are identical to **Eq. 9**. The second term in each of **Eqs. 51** and **52** depends on the input **x**, but this could be replaced with the input drive **z** (where **z** = **W**$_z$ **x**) so as to predict the input drive instead of the input, making it identical to the second term in **Eq. 9**. The last term in each of **Eqs. 51** and **52** is new. This term expresses mutual inhibition between the response of each neuron $y_j$ and the sum of the responses of the other neurons. The neurons compete with one another to encode and predict the input over time, i.e., predictive coding (Harrison 1952; Oliver 1952; Srinivasan et al. 1982).

An example network was constructed to follow an input for past time, but to predict for future time (**Fig. 10**). The input $x(t)$ was a periodic time series, a sum of sinusoids, until $t = 0$ and then nonexistent for $t > 0$ (**Fig. 10a**). The network was constructed with 6 pairs of neurons, corresponding to the real and imaginary parts of $y_j(t)$. The recurrent weight matrix was a diagonal matrix with the real parts of all 6 eigenvalues equal to 1, and with imaginary parts corresponding to 6 different temporal frequency (0, 1, 2, 4, 8, and 16 Hz). Specifically:

$$w_j = 1 + i2\pi\omega_j\tau_y \tag{53}$$

where $w_j$ are the complex-valued weights along the diagonal of the recurrent weight matrix, and $\omega_j$ are the 6 temporal frequencies. This diagonal recurrent weight matrix could, of course, be replaced with a more generic recurrent weight matrix (e.g., analogous to that depicted in **Fig 6a**), with the same complex-valued eigenvalues. The modulators ($a_j$ and $b_j$) were set to the same nonzero value (0.01) at stimulus onset ($t$ = -3000) so that the responses followed the input. Both modulators were set to 0 for $t > 0$ so that the responses continued in spite of the lack of input. Finally, the $a_j$ modulators were set to 1 at $t = 2500$ to reset the responses. The input was real valued (**Fig. 10a**) but the responses were complex valued (**Fig. 10b,c**) because the weights were complex valued. The time-series prediction (**Fig. 10d**, blue curve) was computed by summing the real parts of the responses across the 6 temporal frequencies (i.e., by summing the curves in **Fig. 10b**).

**Figure 10. Time-series prediction. a.** Input, sum of two sinusoids (2 Hz and 8 Hz) for past time ($t \leq 0$) and nonexistent for future time ($t > 0$). **b.** Real parts of the output responses. Each curve corresponds to a different predictive basis-function with 6 different temporal frequencies (0, 1, 2, 4, 8, and 16 Hz). Yellow, 2 Hz. Green, 8 Hz. **c.** Imaginary parts of the output responses. **d.** Summed responses. Blue, real part (sum of curves in panel b). Orange, imaginary part (sum of curves in panel c).

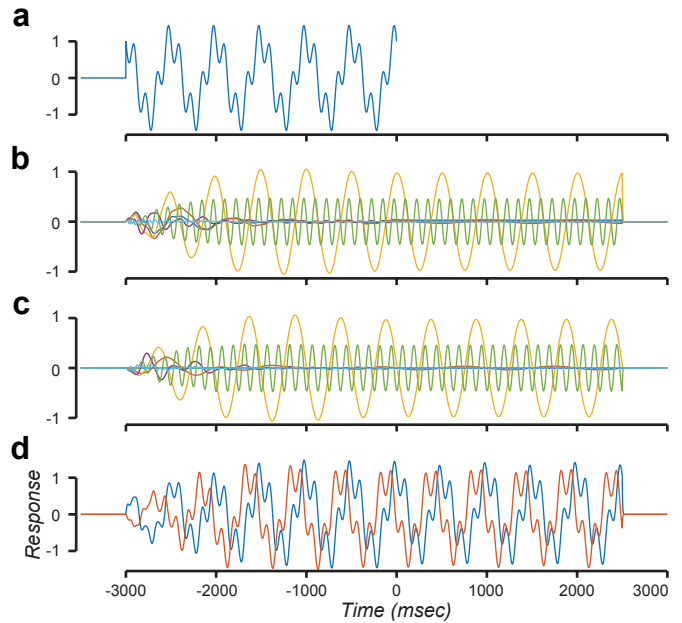

For fixed values of the modulators, each (complex-valued) pair of neurons acts like a shift-invariant linear system (i.e., a recursive linear filter). The predicted responses can be computed recursively, but they can also be expressed as a sum of basis functions that we call the "predictive basis functions". The predictive basis functions (damped oscillators of various temporal frequencies) are the impulse response functions of these shift-invariant linear systems, each corresponding to a different eigenvector/eigenvalue (for a diagonal recurrent weight matrix, each predictive basis function also corresponds to a pair of neurons indexed by $j$). Given the responses of a pair of neurons at only one instant in time, the predicted responses over time are proportional to the predictive basis functions, scaled by the responses at that instant in time. Given the responses over time up to a current instant in time, the predicted responses can be expressed a sum of scaled copies of the predictive basis functions. For the example shown in **Fig. 6**, the first pair of neurons correspond to a temporal frequency of 0 Hz; the predictive basis function is an exponential decay, the response $y_j(t)$ is a low-pass filtered (blurred over time) copy of the input $x(t)$, and the values of the modulators ($a_j$ and $b_j$) determine the amount of blurring.

**Discussion**

ORGaNICs, although introduced as a computational theory of working memory, may be applicable also to models of motor preparation (e.g., as in **Fig. 8**), motor control, sensory processing, prediction over time (e.g., as in **Fig. 10**) and other components of executive control (in addition to working memory, such as cognitive control, e.g., controlling attention). ORGaNICs can be stacked in layers such that the inputs to one ORGaNIC are the outputs from one or more other ORGaNICs. Particular stacked architectures encompass convolutional neural nets (i.e., deep nets) as a special case: specifically when the encoding/embedding weight matrices are convolutional and when the modulators are large ($a_j = b_j \gg 0$) such that the output responses are dominated by the input drive. Consequently, sensory processing (including prediction over time), motor control, and executive control (including working memory along with other cognitive control functions) may all share a common canonical computational foundation.

ORGaNICs build on Heeger's Theory of Cortical Function (TCF) (Heeger 2017) that offers a framework for understanding how the brain accomplishes three key functions: (i) inference: perception is nonconvex optimization that combines sensory input with prior expectation; (ii) exploration: inference relies on neural response variability to explore different possible interpretations; (iii) prediction: inference includes making predictions over a hierarchy of timescales. TCF has a single modulator for all of the neurons in each layer (like $\lambda$ in **Eq. 7**) whereas ORGaNICs may have a separate pair of modulators ($a_j$ and $b_j$) for each neuron. ORGaNICs also have a more general form for the recurrent weight matrix. But TCF includes a feedback drive across the layers of a stacked architecture, in addition to the input drive and recurrent drive. TCF predicts that the brain changes states depending on the values of the modulators. In some states, neural responses are dominated by the feedforward drive and the theory is identical to a conventional feedforward model (e.g., deep net), thereby preserving all of the desirable features of those models. In other states, the theory is a generative model that constructs a sensory representation from an abstract representation, like memory recall. In still other states, the theory combines prior expectation with sensory input, explores different possible perceptual interpretations of ambiguous sensory inputs, and predicts forward in time. The theory, therefore, offers a framework for understanding how the cortex accomplishes inference, ex-ploration, and prediction. TCF is derived from an energy function, very much like **Eq. 1**. Consequently, a variant of ORGaNICs may be stacked (like TCF) to include feedback connections and the capability of a generative model.

*Implications for neuroscience*

1) There is a critical need for developing behavioral tasks that animal models are capable of learning, and that involve both maintaining and manipulating information over time. ORGaNICs (and other LSTMs) manage long-term dependencies between sensory inputs at different times, using a combination of gated integration (e.g., when $b_j > 0$ and $a_j = 0$) and reset (when $a_j > b_j$). Typical delayed-response tasks like the memory-guided saccade task are appropriate for studying what psychologists call "short-term memory", but they are weak probes for studying working memory (Atkinson and Shiffrin 1968; Cowan 1998; Cowan 2008; Postle 2015), because those tasks do not involve manipulation of information over time. Behavioral tasks that are popular in studies of decision making involve integration of noisy sensory information (Brunton et al. 2013; Hanks et al. 2015; Shadlen and Newsome 2001) or integration of probabilistic cues (Yang and Shadlen 2007); variants of these tasks (e.g., Gold and Shadlen 2003; Purcell and Kiani 2016) might be used to test the gated integration and reset functionality of ORGaNICs. The anti-saccade task (Funahashi et al. 1993; Hallett 1978; Johnston and Everling 2006; Munoz and Everling 2004; Saber et al. 2015) and the double-step saccade task (Becker and Jurgens 1979; Goldberg and Bruce 1990; Medendorp et al. 2006; Medendorp et al. 2007; Westheimer 1954) might also be used, with delay periods, to test the theory and to characterize how cortical circuits manage long-term dependencies.

2) Sustained delay-period activity and sequential activation are flip sides of the same coin. ORGaNICs, a straightforward extension of leaky neural integrators, provide a unified theoretical framework for sustained delay-period activity, for sequential activation / synfire chains, and for delay-period activity with complicated dynamics. The responses can be read out at any time during the delay in spite of the complicated dynamics. Indeed, we hypothesize that complicated dynamics is the norm, to support manipulation as well as maintenance.

3) Motor control may share a common computational foundation with sensory processing and working memory, performed with similar circuits. Open-loop (i.e., ballistic) actions can be performed by initializing the state of an ORGaNIC (i.e., with the modulators turned on) from an input, and then by shutting off the modulators to let the dynamics play out. Closed-loop actions may also be performed with the same computation and circuit, in which the input drive reflects sensory feedback during the movement. This idea dovetails with experimental evidence demonstrating that motor cortical preparatory activity functions as advantageous initial conditions for subsequent peri-movement neural dynamics that generate the desired movement (Shenoy et al. 2013). Sensory processing models based on convolutional neural nets (i.e., deep nets), as noted above, are a particular special case of a stacked architecture, with ORGaNICs stacked in layers such that the inputs to one ORGaNIC are the outputs from one or more other ORGaNICs.

4) Stability and E:I balance. The stability of ORGaNICs (and related neural integrator circuits) depends on a combination of the recurrent weight matrix and the relative values of the intrinsic time constants. For example, a differential delay between excitatory and inhibitory neurons can compensate for an imbalance in the excitatory versus inhibitory synaptic weights (**Fig. 7a,b**).

5)  Experiments and experimentally-testable predictions. We posit that the most important contribution of ORGaNICs is the conceptual framework, a canonical computational motif based on gated integration, reset, and controlling the effective time constant. Rethinking cortical computation in these terms should motivate a variety of experiments, some examples of which are as follows. 1) The theory predicts that thalamic input changes the effective time constant and recurrent gain of a PFC neuron. 2) The specific biophysical implementation described above predicts that the soma and basal dendrites share input signals, but with opposite sign. This would, of course, have to be implemented with inhibitory interneurons. 3) The theory predicts that a dynamic time-varying pattern of neural activity (e.g., sequential activity) can nonetheless encode information that is constant over time. In such cases, it ought to be possible to read out the encoded information using a fixed linear decoder in spite of dynamic changes in neural activity. 4) As noted above, variants of sensory integration tasks might be used to test the gated integration and reset functionality of ORGaNICs, and variants of the anti-saccade and double-step saccade tasks might also be used, with delay periods, to characterize how cortical circuits manage long-term dependencies.

_Implications for AI_

1)  Go complex. Simple harmonic motion is everywhere. For many applications (e.g., speech processing, music processing, analyzing human movement), the dynamics of the input signals may be characterized with damped oscillators, in which the amplitudes, frequencies and phases of the oscillators may change over time. The complex-valued weights and responses in ORGaNICs are well-suited for these kinds of signals. Likewise, we propose using damped-oscillator basis functions as a means for predicting forward in time (**Fig. 10**); see also our previous paper outlining a broad theory of cortical function (Heeger 2017). Traditional LSTMs essentially approximate these modulated, oscillatory signals with piecewise constant functions. There has been relatively little focus on complex-valued recurrent neural networks (Fang and Sun 2014; Goh and Mandic 2007; Gong et al. 2015; Hirose 2012; Hu and Wang 2012; Li et al. 2002; Mandic and Goh 2009; Minin et al. 2012; Nitta 2009; Rakkiyappan et al. 2015; Sarroff et al. 2015; Widrow et al. 1975; Zhang et al. 2014; Zhou and Zurada 2009; Zimmermann et al. 2011), and even less on complex-valued LSTMs (Arjovsky et al. 2016; Danihelka et al. 2016; Jing et al. 2017; Jose et al. 2017; Wisdom et al. 2016)

2)  Stability. To ensure stability and to avoid exploding gradients during learning, the recurrent weight matrix should be rescaled so that the eigenvalue with the largest real part is no larger than 1. This rescaling could be added as an extra step during learning after each gradient update. Doing so should help to avoid vanishing gradients by using rectification instead of saturating nonlinearities (see also Jing et al. 2017).

3)  Temporal warping and invariance. A challenge for models of sensory processing is that perception must be invariant with respect to compression or dilation temporal signals (e.g., fast vs. slow speech, Lerner et al. 2014). Likewise, a challenge for models of motor control is to generate movements at different speeds (e.g., playing the same piece of piano music at different tempos). Scaling the time constants of the neurons, all by the same factor, scales the oscillation frequencies by the inverse of that scale factor (**Eqs. 27-28** and **Fig. 7b,c**). This offers a possible solution to the problem of temporal warping (see also Goudar and Buonomano 2018; Gutig and Sompolinsky 2009; Murray and Escola 2017; Tallec and Ollivier 2018). A fixed value for the scale factor would handle linear time-rescaling in which the entire input (and/or output) signal is compressed or dilated by the inverse of the scale factor. It might also be possible to estimate a time-varying value for the scale factor, based on the inputs and/or outputs, to handle time-varying time-warping.

4)   Effective time-constant and recurrent gain. The modulators in ORGaNICs control the effective time constant and recurrent gain. This characterization is complementary to the common description of LSTMs in terms of update gates and reset gates.

5)   Neuromorphic implementation. Given the biophysical (equivalent electrical-circuit) implementation of ORGaNICs, it may be possible to design and fabricate analog VLSI OR-GaNICs chips. Analog circuitry may be more energy-efficient in comparison to representing and processing information digitally (e.g., Mead 1990; Sarpeshkar 1998). Such an analog electrical-circuit may be configured to download various parameter settings (e.g., the weight matrices and offsets), computed separately offline.

## *Learning*

Left open is how to determine the weights in the various weight matrices: the encoding matrix ($\mathbf{W}_{zx}$), the recurrent weight matrix ($\mathbf{W}_{\hat{y}y}$), the readout matrix ($\mathbf{W}_{ry}$), the modulator weight matrices ($\mathbf{W}_{ax}$, $\mathbf{W}_{bx}$, $\mathbf{W}_{ay}$, $\mathbf{W}_{by}$), and the various offsets ($\mathbf{c}_z$, $\mathbf{c}_{\hat{y}}$, $\mathbf{c}_a$, $\mathbf{c}_b$, $\mathbf{c}_r$). A supervised learning approach would estimate the weights via gradient descent, given target values for the response time-courses (or the readout time-courses), if you have access to such target values sampled over time. Another approach would be to learn the weights via adaptive dynamic programming, i.e., the continuous-time equivalent of reinforcement learning (Lewis and Vrabie 2009; Murray et al. 2002; Wang et al. 2009; e.g., Yu et al. 2017). Yet another approach would be an unsupervised learning algorithm based on minimizing prediction error over time (Heeger 2017).

## Appendix: Biophysical implementation

### *Thalamus*

The responses of the modulators (**a** and **b**) were presumed to be computed by neurons in thalamus (Guo et al. 2017; Schmitt et al. 2017). Each neuron was modeled as a single-compartment electrical circuit with conductance-based synapses. We start with the membrane equation:

$$C\frac{dv}{dt} = -g_l(v - E_l) - g_e(v - E_e) - g_i(v - E_i) \quad , \tag{54}$$

The leak conductance, excitatory synaptic conductance, and inhibitory synaptic conductance, are denoted $g_l$, $g_e$, and $g_i$, respectively The corresponding reversal potentials are denoted $E_l$, $E_e$, and $E_i$. To simplify the notation (without loss of generality), we choose $E_l = 0$, $E_e = 1$, and $E_i = -1$. Rewriting **Eq. 54**:

$$C\frac{dv}{dt} = -\left(g_l + g_e + g_i\right)v + g_e - g_i$$
$$\tau\frac{dv}{dt} = -v + \frac{g_e - g_i}{g} \qquad , \tag{55}$$

where we define:

$$g = \left(g_l + g_e + g_i\right) \tag{56}$$

$$\tau = \frac{C}{g} \quad . \tag{57}$$

To compute a linear summation of inputs **x** following by a saturating nonlinearity, we specify the synaptic conductances:

$$g_e = \sum_k w_k^+ x_k^+ + w_k^- x_k^-$$
$$g_i = \sum_k w_k^+ x_k^- + w_k^- x_k^+ \qquad , \tag{58}$$

where $w_k$ are the weights in the weighted sum, and where the superscript + and - mean halfwave rectification:

$$x_k^+ = \lfloor x_k \rfloor \text{ and } x_k^- = \lfloor -x_k \rfloor \quad . \tag{59}$$

Subtracting the two lines of **Eq. 58** gives linear summation:

$$
\begin{aligned}
g_e - g_i &= \sum_k w_k^+ x_k^+ + w_k^- x_k^- - w_k^+ x_k^- - w_k^- x_k^+ \\
&= \sum_k \left(w_k^+ - w_k^-\right)x_k^+ - \left(w_k^+ - w_k^-\right)x_k^- \\
&= \sum_k w_k x_k^+ - w_k x_k^- \\
&= \sum_k w_k x_k
\end{aligned}
\tag{60}
$$

Substituting from **Eq. 60** into **Eq. 55** and solving for the steady state responses gives linear summation followed by a saturating nonlinearity:

$$v = \frac{g_e - g_i}{g} = \frac{1}{g} \sum_k w_k x_k \ . \tag{61}$$

The dynamic, time-varying responses of the full population of thalamic neurons was implemented as follows:

$$\mathbf{x}^+ = \lfloor \mathbf{x} \rfloor \text{ and } \mathbf{x}^- = \lfloor -\mathbf{x} \rfloor$$
$$\mathbf{y}^+ = \lfloor \mathbf{y} \rfloor \text{ and } \mathbf{y}^- = \lfloor -\mathbf{y} \rfloor \tag{62}$$

$$\mathbf{a}^+ = \lfloor \mathbf{a} \rfloor \text{ and } \mathbf{a}^- = \lfloor -\mathbf{a} \rfloor$$
$$\mathbf{W}_{ax}^+ = \lfloor \mathbf{W}_{ax} \rfloor \text{ and } \mathbf{W}_{ax}^- = \lfloor -\mathbf{W}_{ax} \rfloor$$
$$\mathbf{W}_{ay}^+ = \lfloor \mathbf{W}_{ay} \rfloor \text{ and } \mathbf{W}_{ay}^- = \lfloor -\mathbf{W}_{ay} \rfloor \tag{63}$$
$$\mathbf{c}_a^+ = \lfloor \mathbf{c}_a \rfloor \text{ and } \mathbf{c}_a^- = \lfloor -\mathbf{c}_a \rfloor$$

$$\mathbf{b}^+ = \lfloor \mathbf{b} \rfloor \text{ and } \mathbf{b}^- = \lfloor -\mathbf{b} \rfloor$$
$$\mathbf{W}_{bx}^+ = \lfloor \mathbf{W}_{bx} \rfloor \text{ and } \mathbf{W}_{bx}^- = \lfloor -\mathbf{W}_{bx} \rfloor$$
$$\mathbf{W}_{by}^+ = \lfloor \mathbf{W}_{by} \rfloor \text{ and } \mathbf{W}_{by}^- = \lfloor -\mathbf{W}_{by} \rfloor \tag{64}$$
$$\mathbf{c}_b^+ = \lfloor \mathbf{c}_b \rfloor \text{ and } \mathbf{c}_b^- = \lfloor -\mathbf{c}_b \rfloor$$

$$\Delta \mathbf{a}(t) = \frac{\Delta t}{C} \left[ -\mathbf{g}_a(t)\mathbf{a}(t) + \mathbf{g}_{ea}(t) - \mathbf{g}_{ia}(t) \right]$$
$$\mathbf{a}(t + \Delta t) = \mathbf{a}(t) + \Delta \mathbf{a}(t)$$
$$\mathbf{g}_{ea}(t) = \mathbf{W}_{ax}^+ \mathbf{x}^+(t) + \mathbf{W}_{ax}^- \mathbf{x}^-(t) + \mathbf{W}_{ay}^+ \mathbf{y}^+(t) + \mathbf{W}_{ay}^- \mathbf{y}^-(t) + \mathbf{c}_a^+$$
$$\mathbf{g}_{ia}(t) = \mathbf{W}_{ax}^- \mathbf{x}^+(t) + \mathbf{W}_{ax}^+ \mathbf{x}^-(t) + \mathbf{W}_{ay}^- \mathbf{y}^+(t) + \mathbf{W}_{ay}^+ \mathbf{y}^-(t) + \mathbf{c}_a^-$$
$$\mathbf{g}_a(t) = \mathbf{g}_{ea}(t) + \mathbf{g}_{ia}(t) + g_l$$
$$\tag{65}$$

$$\Delta \mathbf{b}(t) = \frac{\Delta t}{C} \left[ -\mathbf{g}_b(t)\mathbf{b}(t) + \mathbf{g}_{eb}(t) - \mathbf{g}_{ib}(t) \right]$$
$$\mathbf{b}(t + \Delta t) = \mathbf{b}(t) + \Delta \mathbf{b}(t)$$
$$\mathbf{g}_{eb}(t) = \mathbf{W}_{bx}^+ \mathbf{x}^+(t) + \mathbf{W}_{bx}^- \mathbf{x}^-(t) + \mathbf{W}_{by}^+ \mathbf{y}^+(t) + \mathbf{W}_{by}^- \mathbf{y}^-(t) + \mathbf{c}_b^+ \quad . \tag{66}$$
$$\mathbf{g}_{ib}(t) = \mathbf{W}_{bx}^- \mathbf{x}^+(t) + \mathbf{W}_{bx}^+ \mathbf{x}^-(t) + \mathbf{W}_{by}^- \mathbf{y}^+(t) + \mathbf{W}_{by}^+ \mathbf{y}^-(t) + \mathbf{c}_b^-$$
$$\mathbf{g}_b(t) = \mathbf{g}_{eb}(t) + \mathbf{g}_{ib}(t) + g_l$$

The dynamic, time-varying responses of the full population of PFC neurons was implemented as follows:

$$\mathbf{y}^+(t) = \lfloor \mathbf{v}^+(t) \rfloor$$
$$\mathbf{y}^-(t) = \lfloor \mathbf{v}^-(t) \rfloor \tag{67}$$

$$\mathbf{W}_{zx}^+ = \lfloor \mathbf{W}_{zx} \rfloor \text{ and } \mathbf{W}_{zx}^- = \lfloor -\mathbf{W}_{zx} \rfloor$$
$$\mathbf{c}_z^+ = \lfloor \mathbf{c}_z \rfloor \text{ and } \mathbf{c}_z^- = \lfloor -\mathbf{c}_z \rfloor \tag{68}$$

$$\mathbf{W}_{\hat{y}y}^+ = \lfloor \mathbf{W}_{\hat{y}y} \rfloor \text{ and } \mathbf{W}_{\hat{y}y}^- = \lfloor -\mathbf{W}_{\hat{y}y} \rfloor$$
$$\mathbf{c}_{\hat{y}}^+ = \lfloor \mathbf{c}_{\hat{y}} \rfloor \text{ and } \mathbf{c}_{\hat{y}}^- = \lfloor -\mathbf{c}_{\hat{y}} \rfloor \tag{69}$$

$$\Delta\mathbf{v}^+(t) = \frac{\Delta t}{C} \left[ -g_{vs}\mathbf{v}^+(t) + \left( \mathbf{I}_z^+(t) - \mathbf{I}_z^-(t) \right) + \mathbf{I}_{as}^+(t) + \mathbf{I}_{bs}^+(t) \right]$$
$$\mathbf{v}^+(t+\Delta t) = \mathbf{v}^+(t) + \Delta\mathbf{v}^+(t)$$
$$\Delta\mathbf{v}_a^+(t) = \frac{\Delta t}{C} \left[ -\mathbf{g}_{va}(t) \odot \mathbf{v}_a^+(t) + \left( \mathbf{I}_{\hat{y}}^+(t) - \mathbf{I}_{\hat{y}}^-(t) \right) - \mathbf{I}_{as}^+(t) \right]$$
$$\mathbf{v}_a^+(t+\Delta t) = \mathbf{v}_a^+(t) + \Delta\mathbf{v}_a^+(t)$$
$$\Delta\mathbf{v}_b^+(t) = \frac{\Delta t}{C} \left[ -\mathbf{g}_{vb}(t) \odot \mathbf{v}_b^+(t) + \left( \mathbf{I}_z^-(t) - \mathbf{I}_z^+(t) \right) - \mathbf{I}_{bs}^+(t) \right]$$
$$\mathbf{v}_b^+(t+\Delta t) = \mathbf{v}_b^+(t) + \Delta\mathbf{v}_b^+(t) \tag{70}$$

$$\Delta\mathbf{v}^-(t) = \frac{\Delta t}{C} \left[ -g_{vs}\mathbf{v}^-(t) + \left( \mathbf{I}_z^-(t) - \mathbf{I}_z^+(t) \right) + \mathbf{I}_{as}^-(t) + \mathbf{I}_{bs}^-(t) \right]$$
$$\mathbf{v}^-(t+\Delta t) = \mathbf{v}^-(t) + \Delta\mathbf{v}^-(t)$$
$$\Delta\mathbf{v}_a^-(t) = \frac{\Delta t}{C} \left[ -\mathbf{g}_{va}(t) \odot \mathbf{v}_a^-(t) + \left( \mathbf{I}_{\hat{y}}^-(t) - \mathbf{I}_{\hat{y}}^+(t) \right) - \mathbf{I}_{as}^-(t) \right]$$
$$\mathbf{v}_a^-(t+\Delta t) = \mathbf{v}_a^-(t) + \Delta\mathbf{v}_a^-(t)$$
$$\Delta\mathbf{v}_b^-(t) = \frac{\Delta t}{C} \left[ -\mathbf{g}_{vb}(t) \odot \mathbf{v}_b^-(t) + \left( \mathbf{I}_z^+(t) - \mathbf{I}_z^-(t) \right) - \mathbf{I}_{bs}^-(t) \right]$$
$$\mathbf{v}_b^-(t+\Delta t) = \mathbf{v}_b^-(t) + \Delta\mathbf{v}_b^-(t) \tag{71}$$

$$\mathbf{g}_{va}(t) = \tfrac{1}{R_a}\mathbf{a}^+(t)$$
$$\mathbf{g}_{vb}(t) = \tfrac{1}{R_b}\mathbf{b}^+(t)$$

(72)

$$\mathbf{I}_z^+(t) = \mathbf{W}_{zx}^+ x^+(t) + \mathbf{W}_{zx}^- x^-(t) + \mathbf{c}_z^+$$
$$\mathbf{I}_z^-(t) = \mathbf{W}_{zx}^- x^+(t) + \mathbf{W}_{zx}^+ x^-(t) + \mathbf{c}_z^-$$

(73)

$$\mathbf{I}_{\tilde{y}}^+(t) = \mathbf{W}_{\tilde{y}y}^+ y^+(t) + \mathbf{W}_{\tilde{y}y}^- y^-(t) + \mathbf{c}_{\tilde{y}}^+$$
$$\mathbf{I}_{\tilde{y}}^-(t) = \mathbf{W}_{\tilde{y}y}^- y^+(t) + \mathbf{W}_{\tilde{y}y}^+ y^-(t) + \mathbf{c}_{\tilde{y}}^-$$

(74)

$$\mathbf{I}_{as}^+(t) = \tfrac{1}{R_a}\left(\mathbf{v}_a^+(t) - \mathbf{v}^+(t)\right)$$
$$\mathbf{I}_{as}^-(t) = \tfrac{1}{R_a}\left(\mathbf{v}_a^-(t) - \mathbf{v}^-(t)\right)$$

(75)

$$\mathbf{I}_{bs}^+(t) = \tfrac{1}{R_b}\left(\mathbf{v}_b^+(t) - \mathbf{v}^+(t)\right)$$
$$\mathbf{I}_{bs}^-(t) = \tfrac{1}{R_b}\left(\mathbf{v}_b^-(t) - \mathbf{v}^-(t)\right)$$

(76)

The circle-dot notation $\odot$ means element-wise multiplication.

The firing rates ($\mathbf{y}^+$ and $\mathbf{y}^-$), synaptic weights ($\mathbf{W}^+$ and $\mathbf{W}^-$) and offsets ($\mathbf{c}^+$ and $\mathbf{c}^-$), shunting conductances ($g_{va}$ and $g_{vb}$), and synaptic currents ($\mathbf{I}_z^+$, $\mathbf{I}_z^-$, $\mathbf{I}_{\tilde{y}}^+$, $\mathbf{I}_{\tilde{y}}^-$) are all non-negative, because of the rectifying nonlinearities. The internal currents ($\mathbf{I}_{as}^+$, $\mathbf{I}_{as}^-$, $\mathbf{I}_{bs}^+$, $\mathbf{I}_{bs}^-$) may be positive or negative, depending on whether current is flowing toward or away from the soma. The minus signs in **Eqs. 70** and **71** may be implemented with inhibitory interneurons that flip the sign by having a hyperpolarizing reversal potential.

And we're done! Phew…

**Acknowledgements**

Funding: None

Special thanks to Mike Halassa, Eero Simoncelli, Mike Landy, Roozbeh Kiani, Charlie Burlingham, Rui Costa, and Walter Senn for comments and discussion.